\documentclass{article}

\usepackage[numbers,sort&compress]{natbib}
\usepackage{amsmath}
\usepackage{enumitem}
\usepackage{graphicx}
\usepackage{multirow, multicol}
\usepackage{adjustbox}
\usepackage{wrapfig}
\usepackage{algorithm} 
\usepackage{algorithmic}
\usepackage{titletoc}
\usepackage{wrapfig}

    \usepackage[final]{neurips_2025}

\usepackage[utf8]{inputenc} 
\usepackage[T1]{fontenc}    
\usepackage{hyperref}       
\usepackage{url}            
\usepackage{booktabs}       
\usepackage{amsfonts}       
\usepackage{nicefrac}       
\usepackage{microtype}      
\usepackage{xcolor}         
\usepackage{xspace}
\usepackage{marvosym}

\hypersetup{
colorlinks=true,
linkcolor=blue,
citecolor=brown
}

\newcommand{\ie}{\emph{i.e.}\xspace}
\newcommand{\eg}{\emph{e.g.}\xspace}

\newcommand{\first}{(\textbf{i})\xspace}
\newcommand{\second}{(\textbf{ii})\xspace}

\newcommand{\pn}{$\pm$}

\newcommand{\T}{^{\textrm T}}
\newcommand{\vct}[1]{\boldsymbol{#1}}

\renewcommand{\paragraph}[1]{\noindent\textbf{#1}\quad}

\newcommand{\framework}{RepLDM\xspace}
\newcommand{\moduleone}{attention guidance\xspace}
\newcommand{\moduleoneshort}{AG\xspace}

\title{\framework: Reprogramming Pretrained Latent Diffusion Models for High-Quality, High-Efficiency, High-Resolution Image Generation}

\author{
  Boyuan Cao\textsuperscript{1}\\
  \And
  Jiaxin Ye\textsuperscript{1}\\
  \And
  Yujie Wei\textsuperscript{1}\\
  \And
  Hongming Shan\textsuperscript{1}\thanks{Corresponding author.}\\
  \and
  $^{1}$Institute of Science and Technology for Brain-Inspired Intelligence \&\\
  MOE Key Laboratory of Computational Neuroscience and Brain-Inspired Intelligence \&\\
  MOE Frontiers Center for Brain Science, Fudan University\\
  \texttt{\{caoby23, jxye22, yjwei22\}@m.fudan.edu.cn, hmshan@fudan.edu.cn}\\
}

\begin{document}

\maketitle

\begin{figure}[h!]
    \vspace{-2.em}
    \centering
    \includegraphics[width=1\textwidth]{figs/teaser.pdf}
    \vspace{-1.8em}
    \caption{\textbf{High-resolution images generated by our \framework using a single consumer-grade 3090 GPU.} The corresponding thumbnails are generated by SDXL~\citep{sdxl} at their training resolution.}
    \label{fig:teaser}
    \vspace{-.5em}
\end{figure}

\begin{abstract}
While latent diffusion models (LDMs), such as Stable Diffusion, are designed for high-resolution (HR) image generation, they often struggle with significant structural distortions when generating images at resolutions higher than their training one.
Instead of relying on extensive retraining, a more resource-efficient approach is to reprogram the pretrained model for HR image generation; however, existing methods often result in poor image quality and long inference time.
We introduce \framework, a novel reprogramming framework for pretrained LDMs that enables \emph{high-quality,  high-efficiency, high-resolution} image generation; see Fig.~\ref{fig:teaser}.
\framework consists of two stages:
(\textbf{i}) an \moduleone stage, which generates a latent representation of a higher-quality training-resolution image using a novel training-free self-attention mechanism to enhance the structural consistency;
and 
(\textbf{ii}) a progressive upsampling stage, which progressively performs upsampling in pixel space to mitigate the severe artifacts caused by latent space upsampling.
The effective initialization from the first stage allows for denoising at higher resolutions with significantly fewer steps, improving the efficiency.
Extensive experimental results demonstrate that \framework significantly outperforms state-of-the-art methods in both quality and efficiency for HR image generation, underscoring its advantages for real-world applications.
Codes: \url{https://github.com/kmittle/RepLDM}.
\end{abstract}

\section{Introduction}
\label{sec:intro}
\begin{figure*}[!t]
    \centering
    \includegraphics[width=1\textwidth]{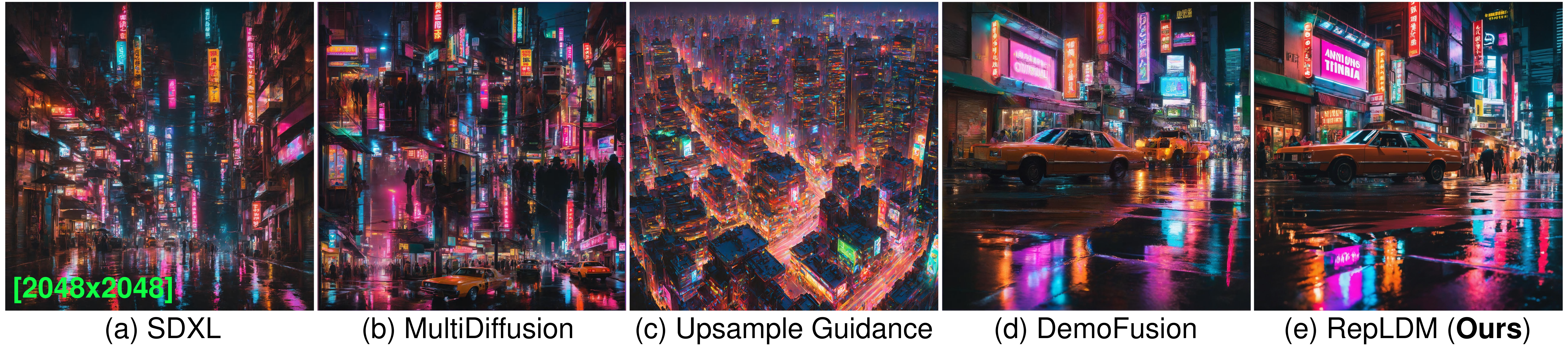}
    \vspace{-1.8em}
    \caption{
    \textbf{Comparison of our \framework with prior work in generating 2048$\times$2048 image}. The prompt is \emph{Neon lights illuminate the bustling cityscape at night, casting colorful reflections on the wet streets}. Zoom-in for a better view.
    }
    \label{fig:models_illustration}
    \vspace{-1em}
\end{figure*}

Diffusion models (DMs) have demonstrated impressive performance in visual generation tasks, particularly in text-to-image generation~\citep{ddpm, iddpm, sdxl, sd3, controlnet, mou2024t2i, ERNIE-ViLG2, wei2024dreamvideo, wei2024dreamvideo2, wei2025dreamrelation, ye2025emotional, wang2024fldm}.
One notable variant of DMs is the latent diffusion model (LDM), which performs diffusion modeling in latent space to reduce training and inference costs, enabling HR generation up to $1024 \times 1024$. 
While it is plausible to modify the input size for higher-resolution generation, this often results in severe structural distortions, as illustrated in Fig.~\ref{fig:models_illustration}(a).
Therefore, a recent research focus is on adapting trained LDMs for HR image generation without the need for additional training or fine-tuning (\ie training-free manner), which can inherit the strong generation capacities of existing LDMs, especially open-sourced versions like Stable Diffusion.

Existing training-free approaches for HR image generation can be roughly categorized into three types: sliding window-based, parameter rectification-based, and progressive upsampling-based.
Sliding window-based methods first divide the HR image into several overlapping patches and use sliding window strategies to perform denoising~\citep{bar2023multidiffusion, haji2023elasticdiffusion, lee2023syncdiffusion}.
However, these methods could result in repeated structures and contents due to the lack of communication between windows; see Fig.~\ref{fig:models_illustration}(b).
Parameter rectification-based methods attempt to correct models' parameters for better structural consistency through the entropy of attention maps,  signal-to-noise ratio, and dilation rates of the convolution layers~\citep{jin2024training, hwang2024upsample, he2023scalecrafter, huang2024fouriscale, zhang2025hidiffusion}. Though efficient, they often lead to the degradation of texture details; see Fig.~\ref{fig:models_illustration}(c).
Unlike the two types mentioned above, progressive upscaling-based methods are to iteratively upsample the image resolution, which maintains better structural consistency and shows state-of-the-art (SOTA) performance~\citep{du2024demofusion, lin2024accdiffusion, lin2024accdiffusionV2, qiu2024freescale}.
Unfortunately, these methods require fully repeating the denoising process multiple times, leading to an unaffordable computational burden; \eg, AccDiffusion~\cite{lin2024accdiffusion} takes 26 minutes to generate a $4096\times 4096$ image. In addition, their upsampling operation in the latent space may introduce artifacts; see Fig.~\ref{fig:models_illustration}(d).
To sum up, existing methods fail to ensure the fast, high-quality HR image generation.

In this paper, we propose \framework, a novel \underline{rep}rogramming framework for pretrained \underline{LDM}s that is capable of generating high-quality, high-resolution images while keeping high-efficiency; see Fig.~\ref{fig:models_illustration}(e).
Specifically, \framework decomposes the denoising process of LDMs into two stages: (\textbf{i}) an \moduleone stage, and (\textbf{ii}) a progressive upsampling stage.
The first stage aims to generate a latent representation of a high-quality image at the training resolution through the proposed \moduleone, which is implemented via a novel training-free self-attention mechanism (TFSA) to improve structural consistency\footnote{
    In this paper, \textit{structural consistency} refers to the plausibility of the overall scene layout and the realism of object structures within an image. Specifically, a reasonable layout should follow logical spatial relationships—for example, the sky should appear above the ground—while realistic object structures should conform to common sense, such as a cat having four legs rather than five.
}.
The second stage aims to progressively upsample the resolution in the pixel space rather than latent space, which can alleviate the severe artifacts caused by the latent space upsampling. 
By leveraging the effective initialization from the first stage, \framework can perform denoising in the second stage with significantly fewer steps, enhancing the overall efficiency with $5\times$ speedup.
Extensive experimental results demonstrate the effectiveness and efficiency of \framework in generating HR images over the SOTA baselines.

The contributions of this work are summarized as follows.
(\textbf{i}) We propose \framework, a novel framework  for high-quality, high-efficiency, high-resolution image generation through  reprogramming pretrained LDMs. 
(\textbf{ii}) We propose \moduleone, which can utilize a novel training-free self-attention to improve the structural consistency of the latent representation towards high-quality images at the training resolution. 
(\textbf{iii}) We propose progressively upsampling the resolution of latent representation in the pixel space, which can alleviate the artifacts caused by the latent space upsampling.
(\textbf{iv}) Extensive experimental results demonstrate that the proposed \framework significantly outperforms the SOTA models in terms of image quality and inference time, emphasizing its great potential for real-world applications.

\section{Related Work}
\label{sec:related}
\paragraph{HR image generation with super-resolution.} An intuitive approach to generating HR images is to first use a pre-trained LDM to generate training-resolution\footnote{
    In this paper, \textit{training resolution} refers to the resolution used during model training, while \textit{high resolution} denotes a resolution that substantially exceeds the training resolution—beyond the level at which the model can directly produce satisfactory results.
}
(TR) images and then apply a super-resolution model to perform upsampling~\citep{wang2023exploiting, zhang2021designing, liang2021swinir, luo2024adaformer, wang2024osffnet}. Although one can obtain structurally consistent HR images in this way, super-resolution models are primarily focused on enlarging the image, and shown to be unable to produce the details that users expect in HR images~\citep{du2024demofusion, lin2024accdiffusion, lin2024accdiffusionV2}.

\paragraph{HR image generation with additional training.} Existing additional training methods either fine-tune existing LDMs with HR images~\citep{hoogeboom2023simple, zheng2024any, guo2024make} or train cascaded diffusion models to gradually synthesize higher-resolution images~\citep{teng2023relay, ho2022cascaded}.
Though effective, these methods require expensive training resources that are unaffordable for regular users.

\paragraph{HR image generation in training-free manner.} Current training-free methods can be roughly classified into three categories: sliding window-based, parameter rectification-based, and progressive upsampling-based methods.
Sliding window-based methods consider spatially splitting HR image generation~\citep{bar2023multidiffusion, haji2023elasticdiffusion, lee2023syncdiffusion}. Specifically, they partition an HR image into several patches with overlap, and then denoise each patch.
However, due to the lack of communication between windows, these methods result in structural disarray and content duplication. While enlarging the overlaps of the windows mitigates this issue, it can result in unbearable computational costs.
For the parameter rectification-based methods, some researchers discovered that the collapse of HR image generation is due to the mismatches between higher resolutions and the model's parameters~\citep{jin2024training, hwang2024upsample, he2023scalecrafter, huang2024fouriscale, zhang2025hidiffusion}.
These methods attempt to eliminate the mismatches by rectifying the parameters such as the dilation rates of some convolutional layers.
While mitigating the structural inconsistency, they often lead to the degradation of image details.
Different from the aforementioned two types, the progressive upsampling-based methods show SOTA performance in some recent studies~\citep{du2024demofusion, lin2024accdiffusion, lin2024accdiffusionV2, qiu2024freescale}.
Though promising, they require fully repeating the denoising process multiple times, which incurs unbearable computational overhead.
Additionally, these methods perform upsampling in the latent space, which may introduce artifacts.

Although their remarkable results, these methods fail to improve the quality of HR images and computational efficiency at the same time.
In contrast, \framework aims to generate HR images with high quality and high efficiency, towards practical applications.

\section{Method}
\label{sec:method}
\subsection{Overview of \framework}
\begin{figure}[t!]
    \centering
    \includegraphics[width=1\textwidth]{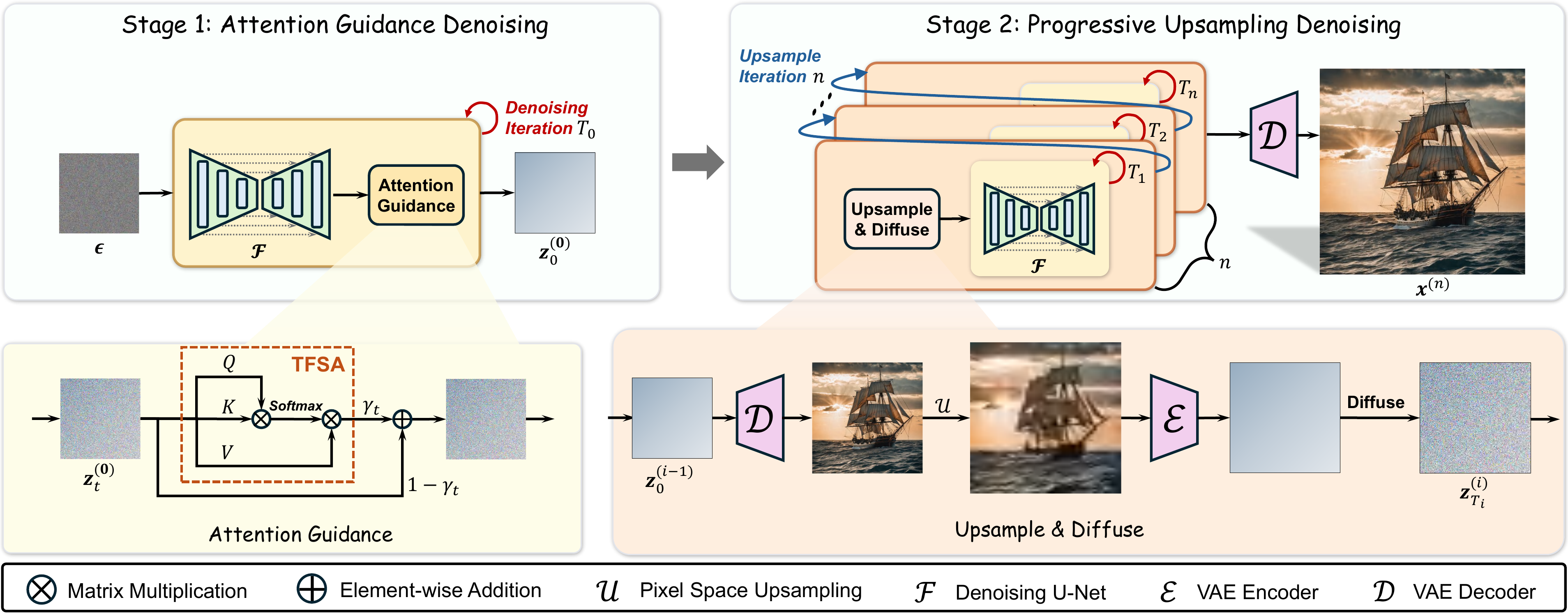}
    \vspace{-1.8em}
    \caption{\textbf{Overview of \framework}. \framework divides the denoising process of a pre-trained LDM into two stages. The first stage leverages the introduced \moduleone to enhance the structural consistency by utilizing a novel training-free self-attention mechanism (TFSA). The second stage iteratively upsamples the latent representation in pixel space to eliminate artifacts.}
    \label{fig:pipeline}
    \vspace{-1em}
\end{figure}

Fig.~\ref{fig:pipeline} presents the overview of \framework, which reprograms a pre-trained LDM to generate HR images without further training. Formally, a pre-trained LDM utilizes a denoising U-Net  model $\mathcal{F}$ to iteratively denoise the latent representation of size  $h\times w \times c$, which is then converted back to the pixel space for final image generation through the decoder $\mathcal{D}$ of a variational autoencoder (VAE). We note that the initial latent representation is sampled from a Gaussian distribution $\boldsymbol{\epsilon}\sim\mathcal{N}(0, \boldsymbol{I})$, and for  inference the encoder $\mathcal{E}$ of VAE is not involved. 

Our \framework extends pre-trained LDMs for higher-resolution image generation in a training-free manner; \ie, $\mathcal{E}$, $\mathcal{D}$ and $\mathcal{F}$ are fixed.
\framework achieves this by decomposing the standard denoising process in the latent space into two stages: (\textbf{i}) \moduleone stage, and (\textbf{ii}) progressive upsampling stage.
In the first stage, \framework aims to generate a latent representation of a higher-quality TR image through the proposed \moduleone.
The \moduleone is implemented as linearly combining the novel training-free self-attention mechanism (TFSA) and original latent representation to improve the structural consistency.
In the second stage, \framework uses the latent representation provided by the first stage as a better initialization, and iteratively obtains higher-resolution images via the pixel space upsampling and diffusion-denoising refinement.

We detail the \moduleone stage in \S\ref{sec:stage1}, followed by the progressive upsampling stage in \S\ref{sec:stage2}.

\subsection{Attention Guidance Stage}
\label{sec:stage1}
\paragraph{Motivation.} Enhancing the structural consistency helps improve image quality~\citep{si2024freeu}. However, it is challenging to do this in a training-free manner.
We observe that the self-attention mechanism presents powerful global spatial modeling capability~\citep{vaswani2017attention, han2021transformer, vit, liu2021swin}, and this capability is parameter-agnostic. It is determined by the paradigm of global similarity calculation inherent to the self-attention mechanism~\citep{vaswani2017attention, zhou2024storydiffusion}.
These insights motivate us to consider designing a novel training-free self-attention mechanism to elegantly enhance the global structural consistency of the latent representation.

\paragraph{Denoising with \moduleone.} To improve the structural consistency of the latent representation at the training resolution $\vct{z}\in\mathbb{R}^{h\times w\times c}$, we propose a simple yet effective training-free self-attention mechanism for \moduleone, termed TFSA, formulated as:
\begin{equation}
\text{TFSA}(\vct{z}) = {f}^{-1}\left(\operatorname{Softmax}\left(\frac{{f}(\vct{z}){f}(\vct{z})\T}{\lambda}\right){f}(\vct{z})\right),
\label{eq:tfsa}
\end{equation}
where the operation ${f}$ reshapes the latent representation into shape $(hw)\times c$ and ${f}^{-1}$ reshapes it back; $\lambda$ is the scaling factor, with a default value of $\lambda=\sqrt{c}$.

However, we empirically observe that directly using the TFSA in Eq.~(\ref{eq:tfsa}) to improve the structural consistency of the latent representation could lead to unstable denoising. Therefore, we propose linearly combining the outputs of TFSA and the original latent representation as \moduleone, which is formulated as:
\begin{equation}
    \tilde{\vct{z}}= \gamma\text{TFSA}\left(\vct{z}\right) + \left(1 - \gamma\right)\vct{z},
\label{eq:attention_guidance}
\end{equation}
where $\tilde{\vct{z}}$ is the structurally enhanced latent representation and $\gamma$ is the guidance scale.
In Appendix~\ref{apx:analyze_ag}, we demonstrate that TFSA functions by modulating the distribution of latent representations.
The term $(1-\gamma)\vct{z}$ in Eq.~\eqref{eq:attention_guidance} serves as a statistical anchor, helping to keep the guided latent representations on the data manifold and ensuring smooth transitions in their distribution.

As shown in Fig.~\ref{fig:pipeline}, we append the \moduleone in Eq.~(\ref{eq:attention_guidance}) to denoising U-Net model $\mathcal{F}$ and repeat the denoising process for a total of $T_0$ times for the first stage. We note that the denoising process starts from step $T_0$ to $1$, and the final output of the first stage is denoted as $\vct{z}_0^{(0)}$.

\paragraph{Adaptive guidance scale.} Considering that the latent representation is mostly non-semantic noise in the first few steps of denoising, we delay $k$ steps in introducing \moduleone.
Moreover, during the denoising process, the image structure is generated first, followed by local details~\citep{yu2023freedom, teng2023relay, luo2024freeenhance}. Therefore, we primarily employ \moduleone in the early to mid-steps of denoising to focus on enhancing the structural consistency of the latent representation. Specifically, we introduce the adaptive guidance scale $\gamma_t$ by applying a decay to a given guidance scale $\gamma$, formulated as:
\begin{equation}
    \gamma_t = 
    \begin{cases}
        \gamma \Big[\frac{1}{2}\big[\cos\left(\tfrac{T_0 - k - t}{T_0 - k} \pi\right) + 1\big] \Big]^\beta & \text{if } t \leq T_0 - k, \\
        0 & \text{otherwise},
    \end{cases}
\end{equation}
where $\beta$ is the decay factor. In practice, considering that $k$ depends on $T_0$ for different resolutions, we use a delay rate $\eta_1=\frac{k}{T_0}$ to control the number of steps for delaying \moduleone.

\subsection{Progressive Upsampling Stage}
\label{sec:stage2}
\begin{figure}[t!]
    \centering
    \includegraphics[width=1\textwidth]{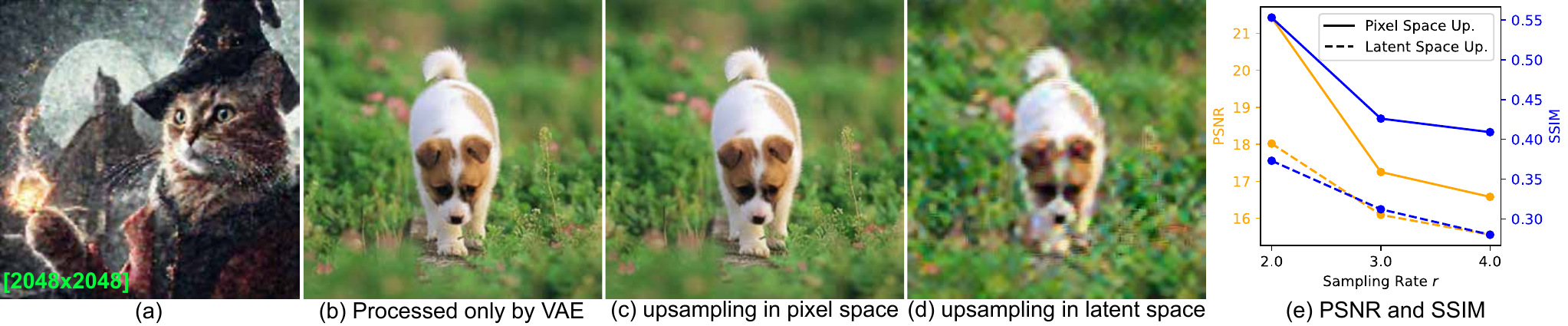}
    \vspace{-1.8em}
    \caption{
    \textbf{Comparison between upsampling in pixel space and latent space.} (a) \framework  with latent space upsampling leads to severe artifacts. (b)-(e): Qualitative and quantitative comparisons of different upsampling methods.
    }
    \label{fig:ablate_interpolation}
    \vspace{-1em}
\end{figure}

\paragraph{Motivation.}  Fig.~\ref{fig:models_illustration}(a) shows that pre-trained LDMs still retain some ability to generate high-frequency information when directly used to synthesize HR images, although they exhibit structural disarray.
Therefore, intuitively, we can utilize the latent representation produced by the first stage as a structural initialization, and generate the HR images  through the ``upsample-diffuse-denoise'' iteration in the latent space.
However, this pipeline leads to severe artifacts, as shown in Fig.~\ref{fig:ablate_interpolation}(a).
We speculate that this is due to \emph{the  upsampling of latent representations in the latent space}.

\paragraph{Pilot study.} To examine this hypothesis, we conduct the following experiments.
Specifically, we randomly select 10k images from ImageNet~\citep{deng2009imagenet} to create an image set $\mathcal{P}$.
For each image $\vct{x} \in \mathcal{P}$, we perform the following operations to obtain three additional image sets:
(\textbf{i}) $\hat{\vct{x}} = \mathcal{D}\circ \mathcal{E}(\vct{x}) $, which use VAE to obtain the reconstructed image set $\mathcal{P}_\text{ref}$;
(\textbf{ii}) $\hat{\vct{x}} = \operatorname{up}\circ\mathcal{D}\circ \mathcal{E} \circ \operatorname{down}(\vct{x}) $, which performs upsampling in pixel space to obtain the image set $\mathcal{P}_\text{pix}$; and 
(\textbf{iii}) $\hat{\vct{x}} = \mathcal{D}\circ \operatorname{up} \circ \mathcal{E} \circ \operatorname{down}(\vct{x})$, which performs upsampling in latent space to obtain the image set $\mathcal{P}_\text{lat}$.
Both upsampling $\operatorname{up}$ and downsampling $\operatorname{down}$ are performed using bicubic interpolation.
Fig.~\ref{fig:ablate_interpolation}(e) reports the quantitative results, where $r$ represents the upsampling or downsampling rate. We calculate the PSNR and SSIM for pixel space upsampling set $\mathcal{P}_\text{pix}$ and latent space upsampling set $\mathcal{P}_\text{lat}$ with respective to the reference set $\mathcal{P}_\text{ref}$.
It can be clearly observed that the latent space upsampling leads to a significant performance decline compared to pixel space upsampling. 
Fig.~\ref{fig:ablate_interpolation}(b-d) shows upsampling in the pixel space produces images close to the reference while upsampling in latent space leads to severe artifacts and detail loss.

\paragraph{Progressive denoising with pixel space upsampling.} Based on the above conclusion, we propose performing upsampling in the pixel space rather than  latent space and utilize diffusion and denoising to refine the upsampled higher-resolution image.
Specifically, the second stage consists of $n$ sub-stages to progressively upsample the training resolution to target resolution, each corresponding to one upsampling operation.
For $i$-th sub-stage, $i=1,\ldots,n$, we prepend an upsample and diffuse operation before the denoising process, which can be defined as:
\vspace{-.5em}
\begin{equation}
\begin{aligned}
    &\hat{\boldsymbol{z}}_0^{(i-1)} = \mathcal{E} \circ \mathcal{U} \circ \mathcal{D}(\boldsymbol{z}_0^{(i-1)}), \\
    &\boldsymbol{z}_{T_i}^{(i)} = \sqrt{\bar{\alpha}_{T_i}} \hat{\boldsymbol{z}}_0^{(i-1)} + \sqrt{1-\bar{\alpha}_{T_i}} \boldsymbol{\epsilon},
\label{eq:noise_inversion}
\end{aligned}
\end{equation}
where $\mathcal{U}$ represents upsampling operation, $\bar{\alpha}_{T_i}$ is the noise schedule hyper-parameter of the $T_i$-th diffusion time step, and $\boldsymbol{z}_0^{(i-1)}$ is the output of the $(i-1)$-th sub-stage; we use  $\boldsymbol{z}_0^{(0)}$ to denote the output from the first stage. Then, $\mathcal{F}$ is used to iteratively denoise $\boldsymbol{z}_{T_i}^{(i)}$ from time step $T_i$ to obtain $\boldsymbol{z}_0^{(i)}$.
After completing all sub-stages, we obtain $z_0^{(n)}$, which is then decoded to produce the final output $\boldsymbol{x}^{(n)} = \mathcal{D}(\boldsymbol{z}_0^{(n)})$.

We empirically found that generating higher-resolution images requires more sub-stages. Additionally, when refining images using diffusion and denoising, higher resolutions demand larger time steps~\citep{teng2023relay}.
In practice, for flexibility, \framework allows users to customize the number of sub-stages $n$, and the diffusion time steps $T_i$ for each sub-stage by a pre-specified  variable-length progressive scheduler $\eta_2 = \left[ \tfrac{T_1}{T_0}, \frac{T_2}{T_0},\ldots, \frac{T_n}{T_0} \right]$. The elements of $\eta_2$ represent the denoising steps of each sub-stage, normalized by $T_0$.

\section{Experiments}
\label{sec:exp}
\subsection{Implementation Details}
\label{sec:exp_implementation}
\paragraph{Experimental settings.}We use SDXL~\citep{sdxl} as the pre-trained LDM and conduct inference using a single NVIDIA 4090 GPU. To ensure consistency when testing inference speed, we use a single 3090 GPU, aligning with other methods.
We randomly sample 33k images from the segment anything model (SAM)~\citep{kirillov2023segment} dataset as the benchmark.
Following the released code from DemoFusion~\citep{du2024demofusion}, we use the EulerDiscreteScheduler~\citep{karras2022elucidating} setting $T_0 = 50$ and the classifier-free guidance~\citep{ho2022classifier} scale to 7.5.
Pixel space upsampling is performed using bicubic interpolation, and the decay factor $\beta$ is fixed at 3.

\paragraph{Evaluation metrics.} The widely recognized metrics Frechet Inception distance (FID)~\citep{heusel2017gans}, Inception score (IS)~\citep{salimans2016improved}, and contrastive language-image pre-training (CLIP) score~\citep{radford2021learning} are used to evaluate model performance.
Additionally, since calculating FID and IS requires resizing images to $299\times 299$, which may not be suitable for evaluating HR images, we follow the experimental settings of~\citep{du2024demofusion, lin2024accdiffusion} to perform ten $1024\times 1024$ window crops on each image to calculate FID$_c$ and IS$_c$.
Since FID is known to be sensitive to small implementation details~\citep{parmar2022aliased}, we employ a widely recognized implementation from a publicly available repository~\citep{Seitzer2020FID}.

\subsection{Quantitative Results}\label{sec:quantitative_exp}
We compare \framework with the following models: (1) SDXL~\citep{sdxl}; (2) MultiDiffusion~\citep{bar2023multidiffusion}; (3) ScaleCrafter~\citep{he2023scalecrafter}; (4) DemoFusion~\citep{du2024demofusion}; (5) Upsample Guidance (UG)~\citep{hwang2024upsample}; (6) AccDiffusion~\citep{lin2024accdiffusion}; and (7) HiDiffusion~\citep{zhang2025hidiffusion}.
For fair comparisons, we disabled the FreeU trick~\citep{si2024freeu} in all experiments.

\begin{table*}[!ht]
\vspace{-1.2em}
\caption{\textbf{Quantitative comparison results}. The best results are marked in \textbf{bold}, and the second best results are marked by \underline{underline}.}
\vspace{-0.em}
\label{table:comparison}
\centering
\small
\begin{adjustbox}{width=1\linewidth,center}
\begin{tabular}{rrrrrrrrrrrrrrrrrrrrr}
\toprule[1.5pt]
\multirow{2}{*}{\textbf{Method}} & \multicolumn{5}{c|}{$2048 \times 2048$} & \multicolumn{5}{c|}{$2048 \times 4096$} & \multicolumn{5}{c|}{$4096 \times 2048$} & \multicolumn{5}{c}{$4096 \times 4096$} \\
\cline{2-21} 
& FID & IS & FID$_{c}$ & IS$_{c}$ & CLIP
& FID & IS & FID$_{c}$ & IS$_{c}$ & CLIP
& FID & IS & FID$_{c}$ & IS$_{c}$ & CLIP
& FID & IS & FID$_{c}$ & IS$_{c}$ & CLIP\\
\midrule[1pt]
SDXL~\citep{sdxl}
& 99.9 & 14.2 & 80.0 & 16.9 & 25.0
& 149.9 & 9.5 & 106.3 & 12.0 & 24.4
& 173.1 & 9.1 & 108.5 & 11.5 & 23.9
& 191.4 & 8.3 & 114.1 & 12.4 & 22.9\\
MultiDiff.~\citep{bar2023multidiffusion}
& 98.8 & 14.5 & 67.9 & 17.1 & 24.6
& 125.8 & 9.6 & 71.9 & \underline{15.7} & \underline{24.6}
& 149.0 & 9.0 & 70.5 & 14.4 & \underline{24.4}
& 168.4 & 6.5 & 76.6 & \underline{14.4} & 23.1\\
ScaleCrafter~\citep{he2023scalecrafter}
& 98.2 & 14.2 & 89.7 & 13.3 & 25.4
& 161.9 & 10.0 & 154.3 & 7.5 & 23.3
& 175.1 & 9.7 & 167.3 & 8.0 & 21.6
& 164.5 & 9.4 & 170.1 & 7.3 & 22.3\\
UG~\cite{hwang2024upsample}
& 82.2 & 17.6 & 65.8 & 14.6 & \textbf{25.5}
& 155.7 & 8.2 & 165.0 & 6.6 & 21.7
& 185.3 & 6.8 & 175.7 & 6.2 & 20.5
& 187.3 & 7.0 & 197.6 & 6.3 & 21.8\\
HiDiff.~\cite{zhang2025hidiffusion}
& 81.0 & 16.8 & 64.1 & 14.2 & 24.9
& 120.7 & 12.2 & 93.0 & 13.6 & 24.2
& 128.4 & 12.8 & 98.3 & 11.3 & 23.1
& 144.1 & 12.5 & 147.0 & 7.4 & 21.2\\
DemoFusion~\cite{du2024demofusion}
& 72.3 & \textbf{21.6} & 53.5 & \textbf{19.1} & \underline{25.2}
& 96.3 & \underline{17.7} & \underline{62.3} & 15.0 & \textbf{25.0}
& \underline{99.6} & \underline{16.4} & \underline{61.9} & \underline{14.7} & \underline{24.4}
& \underline{101.4} & \underline{20.7} & \underline{63.5} & 13.5 & \textbf{24.7}\\
AccDiff.~\citep{lin2024accdiffusion}
& \underline{71.6} & \underline{21.0} & \underline{52.7} & 17.0 & 25.1
& \textbf{95.5} & 16.4 & 62.9 & 11.1 & 24.5
& 102.2 & 15.2 & 65.4 & 11.5 & 24.2
& 103.2 & 20.1 & 65.9 & 13.3 & \underline{24.6}\\
\framework
& \textbf{66.0} & \underline{21.0} & \textbf{47.4} & \underline{17.5} & 25.1
& \textbf{89.0} & \textbf{20.3} & \textbf{56.0} & \textbf{19.0} & \textbf{25.0}
& \textbf{93.2} & \textbf{19.5} & \textbf{56.9} & \textbf{16.5} & \textbf{24.9}
& \textbf{90.6} & \textbf{21.1} & \textbf{59.0} & \textbf{14.8} & \underline{24.6}\\
\bottomrule[1.5pt]
\end{tabular}
\end{adjustbox}
\vspace{-1.em}
\end{table*} 
We report the performance of all methods on four different resolutions (Height $\times$ Width): $4096\times 4096$, $4096\times 2048$, $2048\times 4096$, and $2048\times 2048$.
Considering that the generation time for HR images far exceeds that for low-resolution images, we used 2k prompts at the resolution of $2048\times 2048$, and 1k prompts for resolutions greater $2048\times 2048$.
For all resolutions, we set $\gamma=0.004$, $\beta=3$ and $\eta_1=0.06$ for \framework.
Given that the $4096\times 4096$ resolution is significantly larger than other resolutions, we set $\eta_2 = [0.1, 0.2]$ (\textit{i.e.}, $T_0=50$, $T_1=5$, $T_2=10$) for $4096\times4096$, and $\eta_2=[0.2]$ (\textit{i.e.}, $T_0=50$ and $T_1=10$) for other resolutions.
When generating images with an aspect ratio of $r^{\prime}$, we reshape the initially sampled Gaussian noise $\boldsymbol{\epsilon}$ in the first stage to match $r^{\prime}$.
This process keeps the number of tokens in $\boldsymbol{\epsilon}$ unchanged, preventing drastic fluctuations in the entropy of the attention maps in the transformer~\citep{jin2024training} leading to higher-quality images.

Table~\ref{table:comparison} manifests that \framework significantly outperforms previous SOTA models, AccDiffusion and DemoFusion.
This indicates that \framework generates images with higher quality.
For more comprehensive analyses, we repeat the experiments of Table~\ref{table:comparison} with different random seeds to perform error analyses and conduct a further comparison of the models on the LAION-5B benchmark~\cite{schuhmann2022laion}; see Appendix~\ref{apx:robustness_analysis}.

Table~\ref{table:comparison_time} indicates that \framework demonstrates remarkable advantage in inference speed compared to the SOTA models. On a single 3090 GPU, \framework requires only about one-fifth of the inference time needed by SOTA models such as DemoFusion and AccDiffusion.

\begin{table}[ht!]
\vspace{-.5em}
\caption{\textbf{Model inference time}. The best results are marked in \textbf{bold}. Unit of Time: minute.}
\vspace{-.5em}
\label{table:comparison_time}
\centering
\tiny
\begin{tabular}{ccccccccc}
\toprule[1pt]
\textbf{Resolutions} & SDXL~\citep{sdxl} & MultiDiff.~\citep{bar2023multidiffusion} & ScaleCrafter~\cite{he2023scalecrafter} & UG~\cite{hwang2024upsample} & DemoFusion~\cite{du2024demofusion} & AccDiff.~\citep{lin2024accdiffusion} & HiDiff.~\cite{zhang2025hidiffusion} & \framework\\
\midrule
$2048\times 2048$
& 1.0 & 3.0 & 1.0 & 1.8 & 3.0 & 3.0 & \underline{0.8} & \textbf{0.6}\\
$2048\times 4096$
& 3.0 & 6.0 & 6.0 & 4.0 & 11.0 & 12.7 & \textbf{1.9} & \underline{2.0}\\
$4096\times 4096$
& 8.0 & 15.0 & 19.0 & 11.1 & 25.0 & 26.0 & \textbf{3.4} & \underline{5.7}\\
\bottomrule[1pt]
\end{tabular}
\vspace{-1.em}
\end{table}

\subsection{Qualitative Results}
\label{sec:qualitative_exp}
\begin{figure*}[th!]
    \vspace{-.5em}
    \centering
    \includegraphics[width=1\textwidth]{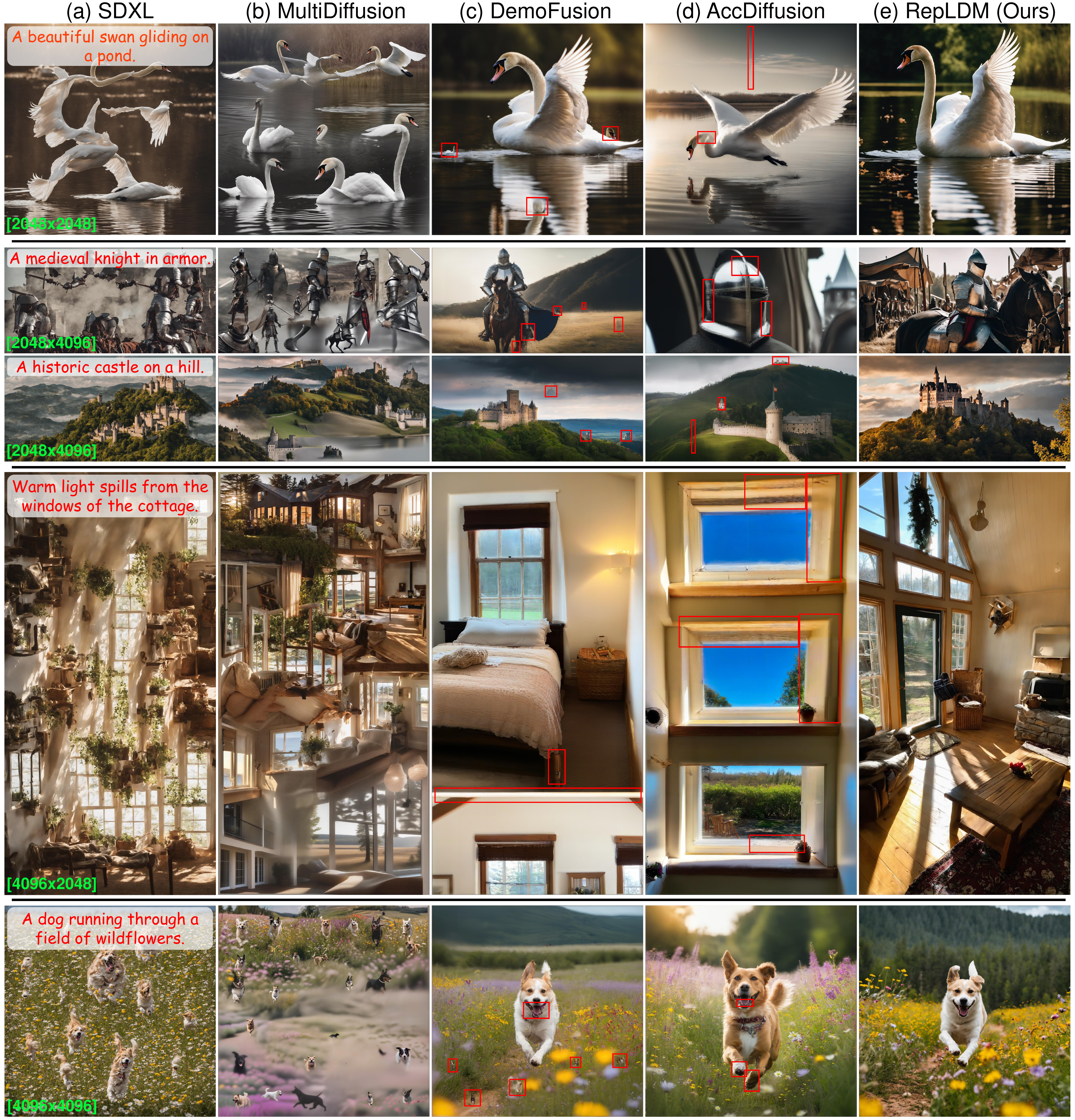}
    \vspace{-1.8em}
    \caption{\textbf{Qualitative comparison with other baselines.} The prompts used to generate the images are presented in the white boxes. MultiDiffusion fails to maintain global semantic consistency. DemoFusion and AccDiffusion exhibit severe artifacts and content repetition. The red boxes indicate some synthesis errors. Zoom-in for a better view.}
    \label{fig:model_comparison}
    \vspace{-1.em}
\end{figure*}

In Fig.~\ref{fig:model_comparison}, \framework is qualitatively compared with AccDiffusion, DemoFusion, and MultiDiffusion. MultiDiffusion fails to maintain global semantic consistency.
As indicated by the red boxes, DemoFusion and AccDiffusion tend to result in chaotic content repetition and severe artifacts, which we speculate are caused by upsampling in the latent space (as analyzed in~\S\ref{sec:stage2}).
In contrast, \framework not only preserves excellent global structural consistency but also synthesizes images with more details.
More qualitative comparison results can be found in Appendix~\ref{apx:supplementary_qualitative_exp}.

\subsection{User Study}
\begin{wraptable}{r}{8.6cm}
\vspace{-2em}
\caption{\textbf{Results of the user study.}}
\begin{adjustbox}{width=1\linewidth,center}
\label{table:user_study}
\centering
\begin{tabular}{r|rr|rr|rr}
\toprule[1pt]
\multirow{2}{*}{\textbf{Method}} & \multicolumn{2}{c|}{Structural Consistency} & \multicolumn{2}{c|}{Color Abundance} & \multicolumn{2}{c}{Detail Richness} \cr\cline{2-7}
& score $\uparrow$ & score* $\uparrow$
& score $\uparrow$ & score* $\uparrow$
& score $\uparrow$ & score* $\uparrow$\\
\midrule
AccDiff.~\citep{lin2024accdiffusion}
& 6.28 & 0.88
& 6.78 & 0.60
& 6.18 & 0.53\\
DemoFusion~\citep{du2024demofusion}
& 5.99 & 0.59
& 6.69 & 0.51
& 6.18 & 0.53\\
\framework
& \textbf{7.42} & \textbf{2.02}
& \textbf{7.64} & \textbf{1.45}
& \textbf{7.41} & \textbf{1.76}\\
\midrule[1pt]
\end{tabular}
\end{adjustbox}
\vspace{-1.2em}
\end{wraptable}

We invite 16 volunteers to participate in a double-blind experiment to further evaluate the performance of the models. Each volunteer is required to answer 35 questions. In each question, three images generated by AccDiffusion, DemoFusion, and \framework are presented. The volunteer needs to rate each image from 1 to 10 in terms of structural consistency, color abundance, and detail richness. We calculate the average of their scores.
Moreover, to eliminate bias in each volunteer's ratings for each metric in each question, we subtract the minimum value among the three scores given by each volunteer for each metric in each question. The rectified score is denoted as score*.
Table~\ref{table:user_study} shows that \framework surpasses previous SOTA models across all metrics.

\section{Ablation Study}
\label{sec:ablation}

\subsection{Attention Guidance}
\label{sec:ag_ablation}
In this section, we first conduct ablation experiments on \moduleone, followed by ablation experiments on the hyper-parameters of \moduleone.
In Appendix~\ref{apx:analyze_ag}, we provide a detailed analysis of how \moduleone improves latent structural consistency and image quality.

\paragraph{Ablation on \moduleone.} We keep $\eta_2$ unchanged and analyze the effect of \moduleone through qualitative and quantitative experiments.
Table~\ref{table:ablation_ag} shows that \moduleone leads to improvements across various metrics, indicating that using \moduleone to enhance the consistency of latent encoding results in higher-quality images.
The qualitative experiments in Fig.~\ref{fig:ablate_ag} demonstrate that using \moduleone eliminates image blurriness and enriches the image details.
Note that FID and IS quantify the statistical differences between two distributions~\cite{barratt2018note, salimans2016improved, heusel2017gans}.
Since \moduleone mainly enhances visual quality by modifying the mid- and high-frequency components while preserving the low-frequency structure of the image, it has limited impact on the overall distributional statistics.
Although \moduleone may not yield significant improvements in quantitative metrics, it provides a noticeable enhancement in human visual perception; see Table~\ref{table:user_study}.
Please refer to Appendix~\ref{apx:supplementary_ablation_ag} for additional qualitative ablation results.
\begin{table*}[ht!]
\vspace{-1em}
\caption{\textbf{Ablation on \moduleone (\moduleoneshort).} The best results are marked in \textbf{bold}.}
\label{table:ablation_ag}
\centering
\begin{adjustbox}{width=1\linewidth,center}
\begin{tabular}{rrrrrrrrrrrrrrrrrrrrr}
\toprule[1.5pt]
\multirow{2}{*}{\textbf{Method}} & \multicolumn{5}{c|}{$2048 \times 2048$} & \multicolumn{5}{c|}{$2048 \times 4096$} & \multicolumn{5}{c|}{$4096 \times 2048$} & \multicolumn{5}{c}{$4096 \times 4096$} \cr\cline{2-21} 
& FID & IS & FID$_{c}$ & IS$_{c}$ & CLIP
& FID & IS & FID$_{c}$ & IS$_{c}$ & CLIP
& FID & IS & FID$_{c}$ & IS$_{c}$ & CLIP
& FID & IS & FID$_{c}$ & IS$_{c}$ & CLIP\\
\midrule[1pt]
w/o \moduleoneshort
& 66.8 & \textbf{21.6} & 47.5 & 17.4 & \textbf{25.3}
& 91.6 & \textbf{20.3} & 58.0 & 14.5 & \textbf{25.0}
& 95.3 & \textbf{19.9} & 58.4 & 14.5 & \textbf{24.9}
& 92.0 & \textbf{21.6} & 59.8 & 13.6 & 24.5\\
w/{\hspace{2mm}} \moduleoneshort
& \textbf{66.0} & 21.0 & \textbf{47.4} & \textbf{17.5} & 25.1
& \textbf{89.0} & \textbf{20.3} & \textbf{56.0} & \textbf{19.0} & \textbf{25.0}
& \textbf{93.2} & 19.5 & \textbf{56.9} & \textbf{16.5} & \textbf{24.9}
& \textbf{90.6} & 21.1 & \textbf{59.0} & \textbf{14.8} & \textbf{24.6}\\
\bottomrule[1.5pt]
\end{tabular}
\end{adjustbox}
\vspace{-1em}
\end{table*}
\begin{figure}[!th]
    \vspace{-.5em}
    \centering
    \includegraphics[width=1\textwidth]{figs/ablation_ag.pdf}
    \vspace{-2.em}
    \caption{\textbf{Ablation on \moduleone}. Zoom-in for a better view.}
    \label{fig:ablate_ag}
    \vspace{-1.5em}
\end{figure}

\begin{figure*}[!th]
    \centering
    \includegraphics[width=1\textwidth]{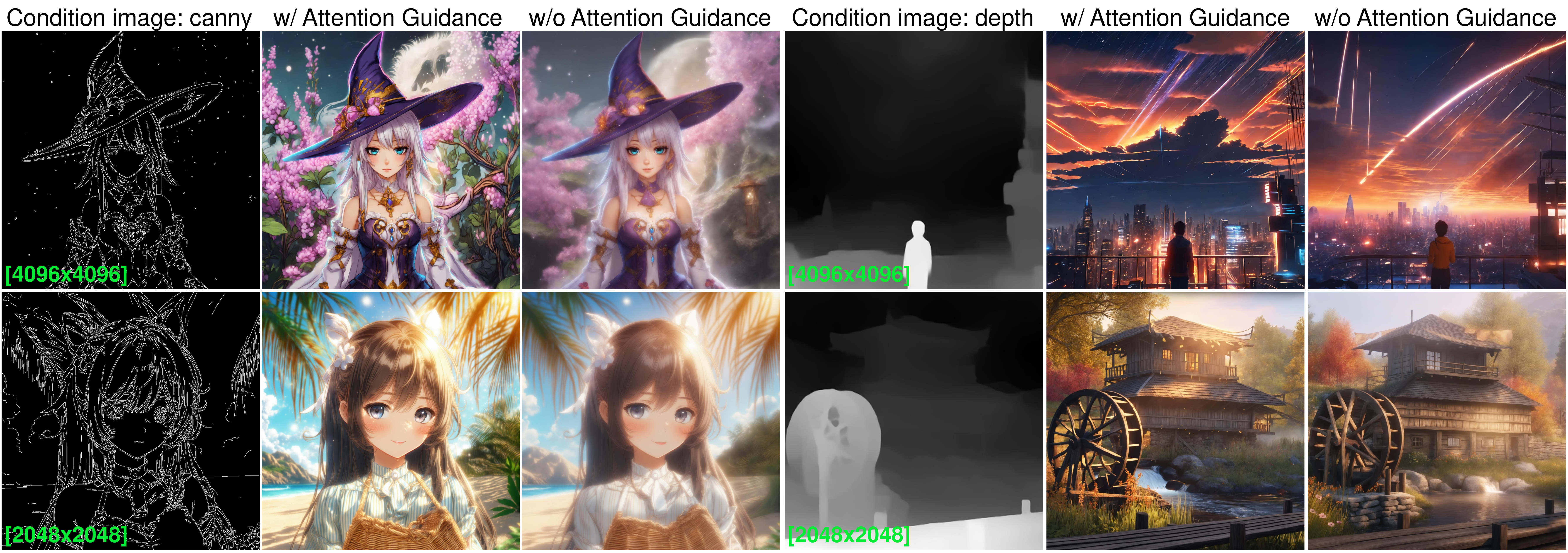}
    \vspace{-2em}
    \caption{\textbf{Ablation on \moduleone with ControlNet.}}
    \vspace{-1.5em}
    \label{fig:ablation_controlnet}
\end{figure*}

\paragraph{Ablation on \moduleone with ControlNet.}
To further demonstrate the generalization ability of \moduleone, in this section, we perform an qualitative ablation study of \moduleone with ControlNet~\cite{controlnet}.
Specifically, we conducted comparative experiments using two types of conditional guidance (canny and depth) across two resolution scales: $4096\times 4096$ and $2048\times 2048$.
As shown in Fig.~\ref{fig:ablation_controlnet}, the integration of \moduleone with ControlNet substantially enhances chromatic fidelity and structural granularity in synthesized images.

\begin{figure}[!th]
    \vspace{-.5em}
    \centering
    \includegraphics[width=1\textwidth]{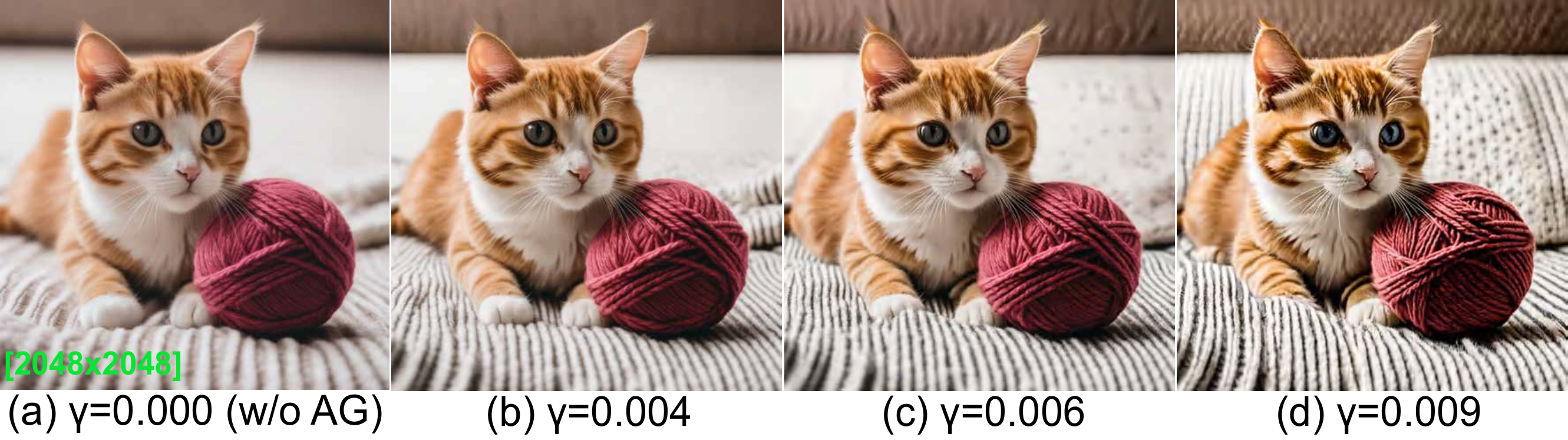}
    \vspace{-2.em}
    \caption{\textbf{Ablation on guidance scale}. Zoom-in for a better view.}
    \label{fig:ablate_guidance_scale}
    \vspace{-1.2em}
\end{figure}

\paragraph{Ablation on guidance scale $\gamma$.} We fix $\eta_1=0.06, \eta_2=[0.2]$ and then explore the effect of the guidance scale $\gamma$ through both quantitative and qualitative experiments.
For the quantitative experiments, we find that $\gamma=0.004$ performs better.
Interestingly, when a larger $\gamma$ is used, the visual quality of the images can be further enhanced.
As shown in Fig.~\ref{fig:ablate_guidance_scale}, using a larger guidance scale results in richer image details. This allows users to generate images according to their preferences for detail richness and color contrast by adjusting the guidance scale.
The setup and results of the quantitative experiments are detailed in Appendix~\ref{apx:supplementary_ablation_ag}.

\begin{wrapfigure}{r}{8.5cm}
    \vspace{-1.1em}
    \centering
    \includegraphics[width=0.60\textwidth]{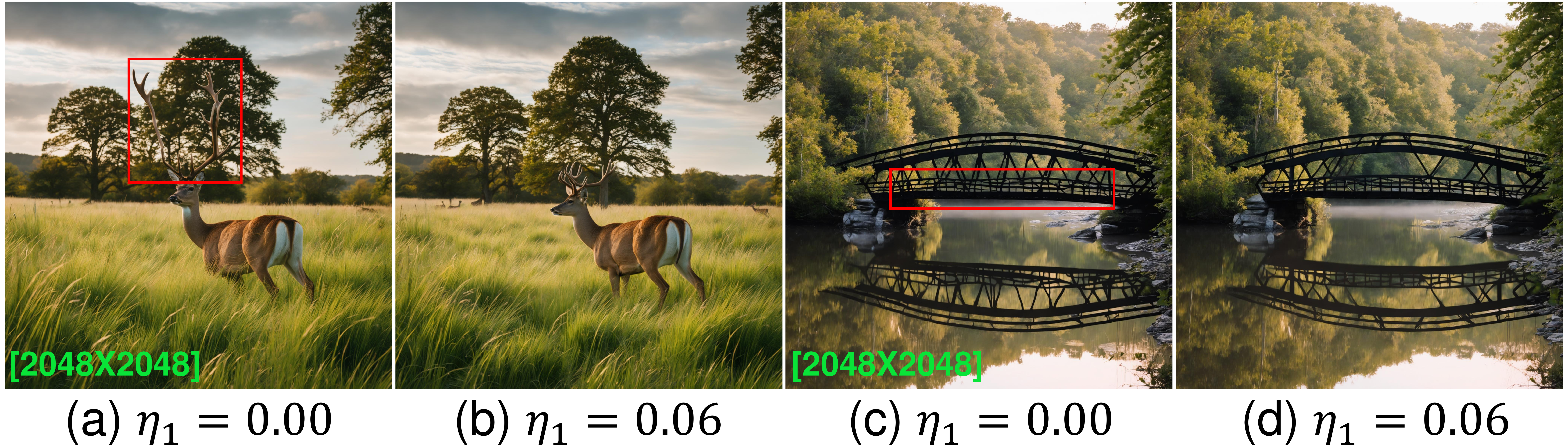}
    \vspace{-1em}
    \caption{\textbf{Ablation on delay rate.} Errors indicated by red boxes can be eliminated by delaying \moduleone. Zoom-in for a better view.}
    \vspace{-1em}
    \label{fig:ablate_delay_rate}
\end{wrapfigure}
\paragraph{Ablation on delay rate $\eta_1$.} We fix $\gamma=0.004, \eta_2=[0.2]$ and then investigate the impact of the delay rate $\eta_1$ through both quantitative and qualitative experiments.
The quantitative analysis results indicate that better generation results can be achieved when $\eta_1 = 0.06$, indicating that appropriately delaying the effect of \moduleone contributes to further improving the quality of the images.
We conjecture that this is because, at the very beginning of the denoising process, the structural information in the latent encoding has not yet emerged, and thus \moduleone cannot effectively enhance structural consistency.
As shown in Fig.~\ref{fig:ablate_delay_rate}, delaying the effect of \moduleone eliminates some generation errors, further improving image quality.
The setup and results of the quantitative experiments are detailed in Appendix~\ref{apx:supplementary_ablation_ag}.

\begin{wrapfigure}{r}{8.5cm}
    \vspace{-1.1em}
    \centering
    \includegraphics[width=0.60\textwidth]{figs/ablation_ag_stage.pdf}
    \vspace{-1em}
    \caption{\textbf{Applying \moduleone at different denoising steps.} Zoom-in for a better view.}
    \vspace{-1em}
    \label{fig:ablate_ag_stage}
\end{wrapfigure}
\paragraph{Ablation on the time steps of \moduleone.} To explain why \moduleone needs to be applied during the early to middle steps of denoising, we apply \moduleone during different denoising steps of the first stage: (a) 47 to 33, (b) 32 to 17, and (c) 16 to 1.
Fig.~\ref{fig:ablate_ag_stage} shows that when \moduleone is applied during the early to middle steps of denoising, the image becomes clearer and more detailed; however, when \moduleone is applied during the later steps of denoising, it has negligible effect on the generated image.
We speculate that this is because diffusion models tend to synthesize structural information first~\citep{teng2023relay, yu2023freedom, luo2024freeenhance}, and once the structural information is generated, \moduleone may have a limited impact on structural consistency.

\subsection{Progressive High-Resolution Denoising}
\label{sec:progressive_ablation}

\begin{wrapfigure}{r}{8.5cm}
    \vspace{-1.2em}
    \centering
    \includegraphics[width=0.60\textwidth]{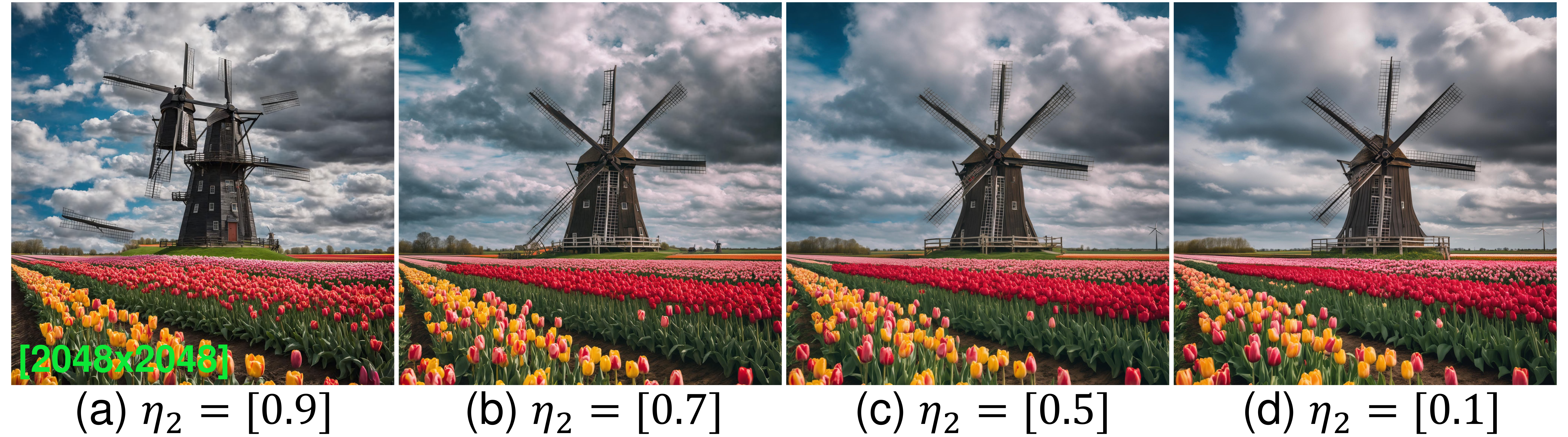}
    \vspace{-1em}
    \caption{\textbf{Generated images using different $\eta_2$.} (a): When the value of progressive scheduler is too large, the structural repetition issue may reappear. (b) to (d): The visual effects are similar. Therefore, we can use a smaller progressive scheduler value to accelerate inference.}
    \vspace{-1em}
    \label{fig:ablation_init_rate}
\end{wrapfigure}
In this section, we conduct ablation experiments on the progressive scheduler $\eta_2$ in the second stage of \framework.
Specifically, we fixed $\gamma=0, \eta_1=0$ and then explore the effect of the progressive scheduler $\eta_2$ through both quantitative and qualitative experiments.
Quantitative experimental results indicate that an excessively large progressive scheduler value may result in a decline in image quality.
This can also be observed in Fig.~\ref{fig:ablation_init_rate}. It is evident that a too large progressive scheduler value may lead to structural misalignment and repetition issues observed in pre-trained SDXL. When the progressive scheduler value is sufficiently small, changing it yields similar visual effects. Therefore, we can choose a smaller progressive scheduler value (\eg, 0.2) to accelerate inference.
The setup and quantitative results are detailed in Appendix~\ref{apx:supplementary_ablation_ag}.

\section{Limitations And Future Work}
\label{sec:limitations}

\begin{wrapfigure}{r}{8.5cm}
\vspace{-1.2em}
\centering
\includegraphics[width=0.60\textwidth]{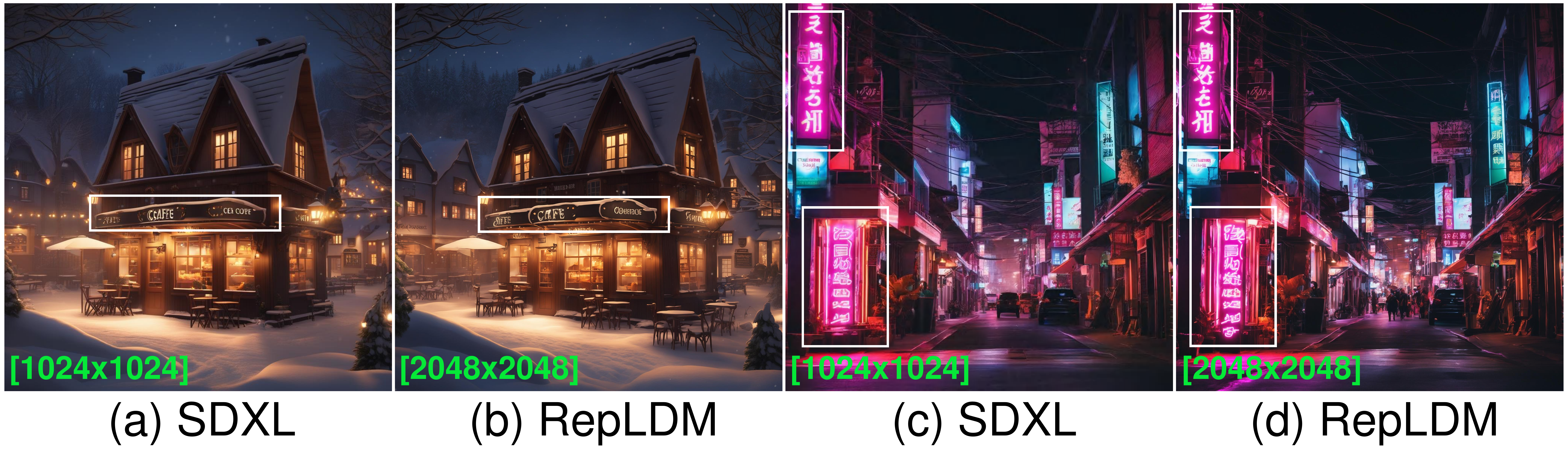}
\vspace{-1em}
\caption{\textbf{Limitations of \framework.} The generation results of SDXL at its training resolution and those of \framework at higher resolutions are provided. As indicated by the white boxes, \framework fails to address the text structure errors inherited from SDXL.}
\vspace{-1em}
\label{fig:limitations}
\end{wrapfigure}
\framework exhibits limitations in the following aspects: (\textbf{i}) Effectively controlling text in images is challenging, as demonstrated by examples in Fig.~\ref{fig:limitations}.
This may be due to the inherent limitations of SDXL in generating textual symbols.
Text, due to its more regular structure compared to other image content, is difficult to restore by directly enhancing the structural consistency of the latent representation.
We speculate that the most reliable approach would be to fine-tune the model specifically on images containing text.
(\textbf{ii}) When generating ultra-high resolution images, such as $12800\times 12800$, the second stage of \framework inevitably needs to be decomposed into more sub-stages, which increases the model's inference time.

Developing a low-cost and effective fine-tuning method to correct text generation errors may be a promising topic.
Moreover, adapting \moduleone to other tasks, such as video generation can be an interesting issue.

\section{Conclusion}
\label{sec:conclusion}

In this paper, we reprogram pretrained LDMs, unlock their potentials, and propose \framework for high-quality, high-efficiency, high-resolution image generation.
\framework divides the denoising process of an LDM into two stages: (\textbf{i}) \moduleone stage, and (\textbf{ii}) progressive upsampling stage.
The first stage generates structurally enhanced latent representations through the proposed \moduleone, employing a novel parameter-free self-attention mechanism.
The second stage iteratively performs upsampling in the pixel stage, thus eliminating the artifacts caused by latent space upsampling.
Extensive experiments show that our proposed \framework significantly outperforms SOTA models while achieving $5\times$ speedup in HR image generation.

\section*{Acknowledgement}
This work was supported by the National Natural Science Foundation of China (No. 62471148) and the computations in this research were supported by the CFFF platform of Fudan University.

\bibliographystyle{plain}
\bibliography{references}

\newpage
\appendix

\startcontents[sections]
\printcontents[sections]{}{1}{\section*{Appendix}}

\section{How Does TFSA/Attention Guidance Work?}
\label{apx:analyze_ag}

In this section, we further elaborate on the working mechanism of \moduleone. Our \moduleone enhances the structural consistency of the latent representation by integrating the output of TFSA. Therefore, we conduct a detailed analysis of TFSA.
Specifically, the functionality of TFSA can be described in two aspects: (\textbf{i}) \textit{clustering the related tokens} in the latent representations;
(\textbf{ii}) \textit{adjusting the amplitude of the high-frequency and low-frequency components} in the latent representations.

\subsection{TFSA Clusters Semantically Related Tokens}

\begin{figure*}[!ht]
    \centering
    \includegraphics[width=1\textwidth]{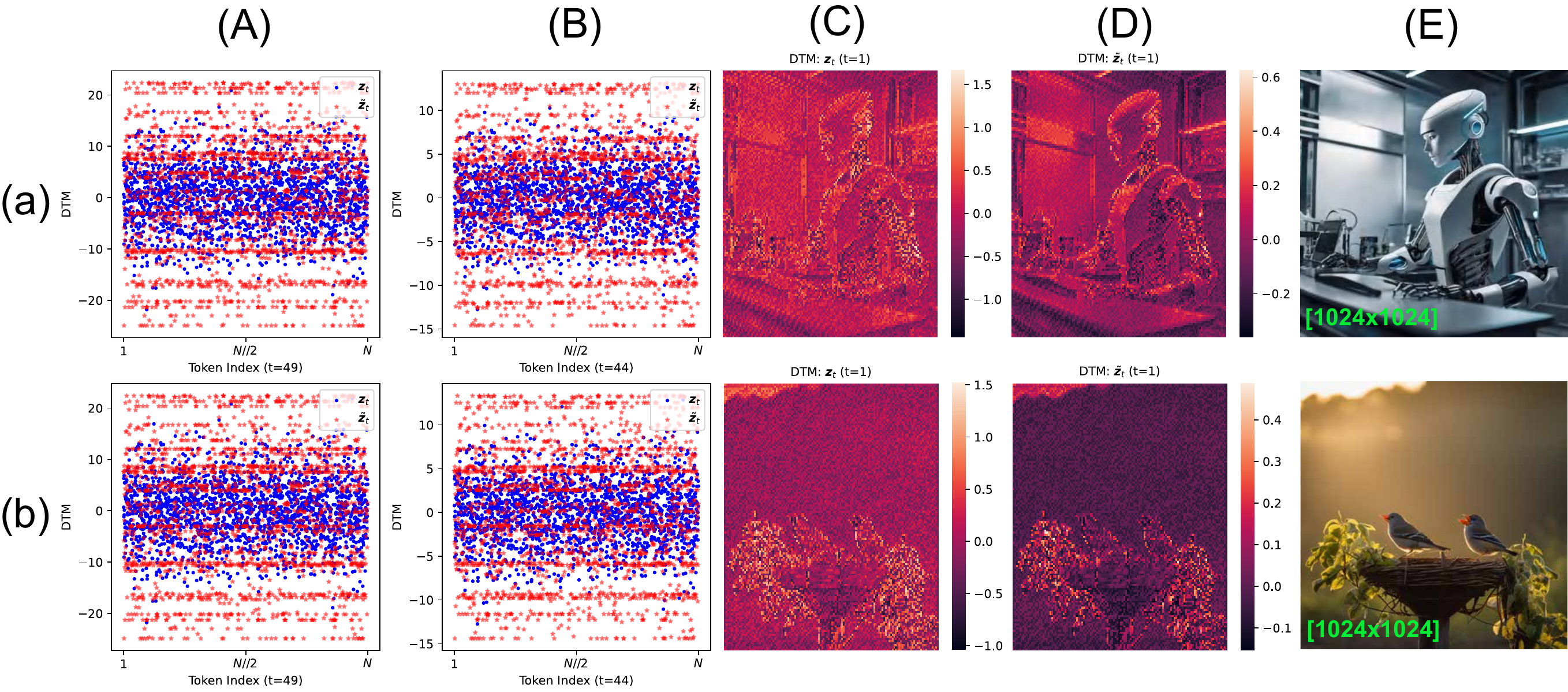}
    \vspace{-2em}
    \caption{
        \textbf{The clustering effect of TFSA.}
        Columns (A), (B), (C), and (D) show the DTM of latent representations, while column (E) presents the corresponding generated RGB images.
    }
    \label{fig:cluster}
    \vspace{-1em}
\end{figure*}

\paragraph{Visualization of the clustering effect of TFSA.}
TFSA reorganizes tokens based on their similarities.
Intuitively, this enables TFSA to perform token clustering, which enhances the structural consistency of latent representations.
To demonstrate the clustering effect of TFSA, we calculated the deviation of the tokens' mean (DTM) of the latent representations $\tilde{\vct{z}}_t$ and $\vct{z}_t$. Concretely, assuming $\vct{z}_t \in \mathbb{R}^{h\times w\times c}$, and $\vct{Z}_t=\operatorname{Flatten}(\vct{z}_t)=[\vct{y}_{t1},\ldots,\vct{y}_{tN}] \in \mathbb{R}^{N\times c}$, where $N=h\times w$, we calculate DTM as:
\begin{equation}
    \text{DTM} = [\operatorname{mean}(\vct{y}_{ti}) - \operatorname{mean}(\vct{Z}_t) \ \text{for}\  i=1,\ldots,N]
\end{equation}
To provide an intuitive illustration of the clustering effect of TFSA, we visualize the DTM based on token indices (\ie, \(i = 1, \dots, N\)) when \(t\) is relatively large.
As shown in columns (A) and (B) of Fig.~\ref{fig:cluster}, compared to the DTM of \(\vct{z}_t\) (blue points), the DTM of \(\tilde{\vct{z}}_t\) (red points) becomes more dispersed and exhibits distinct stripe patterns, indicating that TFSA indeed clusters the tokens of the latent representations.
This clustering effect can be more directly demonstrated when \(t\) is smaller. As shown in the heatmaps in columns (C) and (D) of Fig.~\ref{fig:cluster}, it is evident that TFSA clusters semantically related tokens.

\begin{figure}[!t]
    \centering
    \includegraphics[width=1\linewidth]{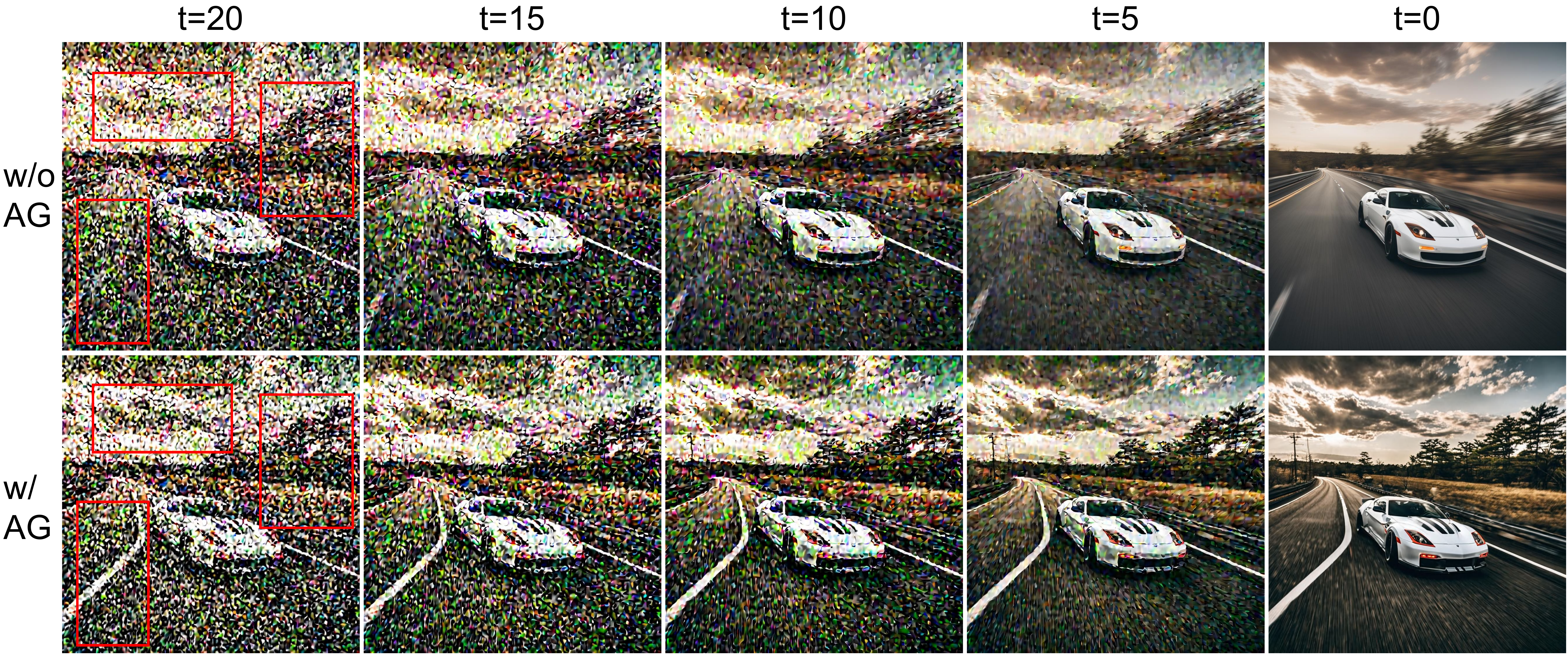}
    \vspace{-2.em}
    \caption{
        \textbf{Denoising visualization for the ablation of \moduleone.}
        As indicated by the red boxes, the clustering effect of TFSA prompts earlier structural emergence, delivering better prior for subsequent fine-detail generation.
        Resolution: $1024\times 1024$.
    }
    \label{fig:visulize_denosing}
    \vspace{-.75em}
\end{figure}

\paragraph{The clustering effect of TFSA leads to accelerated structural denoising.}
Fig.~\ref{fig:cluster} shows that the clustering effect of TFSA clarifies the semantic structures of objects, enabling the model to complete the denoising of low-frequency structures earlier.
This early revelation of the overall image layout provides a stronger prior for subsequent fine-detail generation.
To illustrate this, Fig.~\ref{fig:visulize_denosing} presents the denoising process for the ablation of \moduleone.
Note the regions highlighted by red boxes. With the incorporation of \moduleone, these areas exhibit clearer structures, which facilitates the generation of more affluent details and more vivid colors in subsequent steps.

\begin{wrapfigure}{r}{7.0cm}
    \vspace{-1.4em}
    \centering
    \includegraphics[width=1\linewidth]{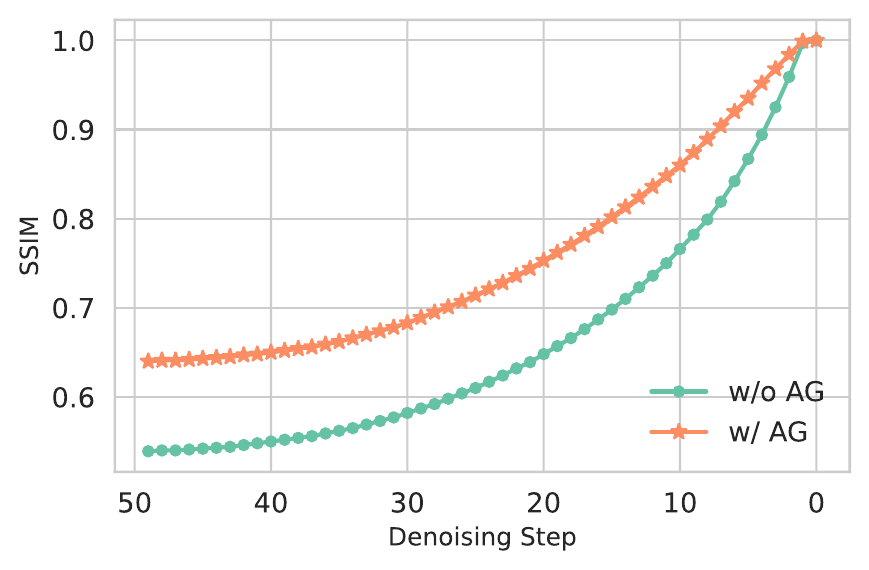}
    \vspace{-2.3em}
    \caption{\textbf{Quantitatively analysis on the clustering effect of TFSA.} We calculate the SSIM between noised latents $\vct{z}_t$ ($1\leq t \leq 49$) and their corresponding clean latent $\vct{z}_0$.}
    \vspace{-3em}
    \label{fig:ssim_latent}
\end{wrapfigure}
To quantitatively demonstrate that TFSA accelerates structural emergence, we calculate the SSIM between $\vct{z}_t$ and $\vct{z}_0$, where $t\in{1,2\ldots,T-1}$, and $T=50$.
As shown in Fig.~\ref{fig:ssim_latent}, compared to the naive denoising process, \moduleone consistently drives the latent representations closer to their final states at each step, indicating the structural foreseeability of TFSA.

\subsection{TFSA Adjusts the Amplitude of High- and Low-frequency Components}

\begin{figure*}[!ht]
    \centering
    \includegraphics[width=1\textwidth]{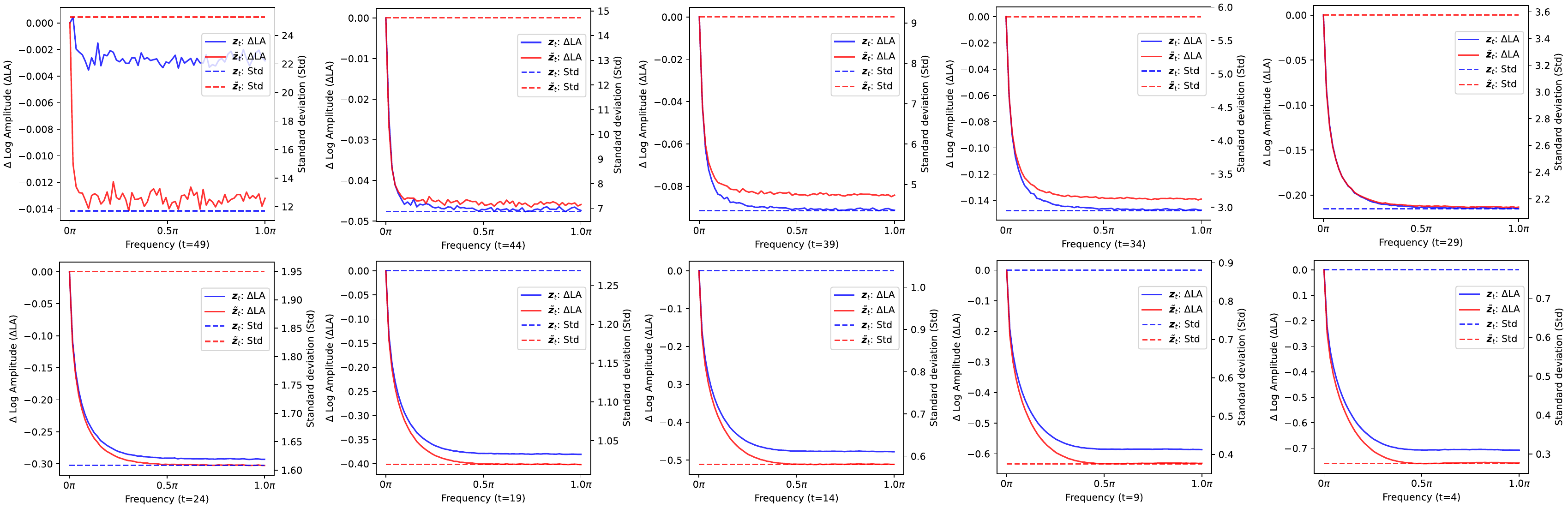}
    \vspace{-2em}
    \caption{
        \textbf{The Fourier transform of the latent representation and the mean of the standard deviations across all channels.}
        \(\vct{z}_t\) is represented in blue, while \(\tilde{\vct{z}_t}\) is represented in red; the Fourier transforms are shown as solid lines, and the standard deviations are shown as dashed lines.
        The results are based on the generation process of 5k images.
    }
    \label{fig:amp-freq}
    \vspace{-1.5em}
\end{figure*}

The aim of this experiment is to explain: (\textbf{i}) why appropriately delaying \moduleone can resolve structural deformation issues (as shown in Fig.~\ref{fig:ablate_delay_rate}); (\textbf{ii}) why \moduleone enhances the details and colors of the image (as shown in Fig.~\ref{fig:ablate_ag} and \ref{fig:ablate_guidance_scale}); and (\textbf{iii}) why applying \moduleone in the later stages of denoising does not enhance the image details and colors (as shown in Fig.~\ref{fig:ablate_ag_stage}).

To explain the aforementioned three points, as shown in Fig.~\ref{fig:amp-freq}, we calculate the Fourier transforms of $\vct{z}_t$ (blue solid line) and $\tilde{\vct{z}}_t$ (red solid line), along with the mean of the standard deviations for all their channels (dashed line). 
It can be observed that TFSA significantly alters the relative amplitudes of the high- and low-frequency components in the latent representations during the initial denoising steps (from \(t = 49\) to \(t = 47\)), particularly affecting the low-frequency components, which results in structural deformation.
During the early and middle stages of denoising (from \(t = 44\) to \(t = 29\)), TFSA increases the amplitudes of high-frequency components in the latent representations, which explains why \moduleone leads to richer details and colors.
In the later stages of denoising (from \(t = 28\) to \(t = 0\)), TFSA slightly suppresses the high-frequency components of the latent representations while almost leaving the low-frequency components unchanged. This explains why applying \moduleone in the later stages of denoising cannot enrich details and colors of the generated images.

Additionally, Fig.~\ref{fig:amp-freq} shows that TFSA increases the standard deviation of $\tilde{\vct{z}}_t$ during the early and middle stages of denoising, while decreasing it in the later stages. The trend of the standard deviation changes is closely consistent with the variation in the amplitude of the high-frequency components. We conjecture that this is because the amount of information in the latent representations is positively correlated with the standard deviation, where a larger standard deviation corresponds to more image details and larger high-frequency components.

\begin{figure}[!t]
    \vspace{-.5em}
    \centering
    \includegraphics[width=1\linewidth]{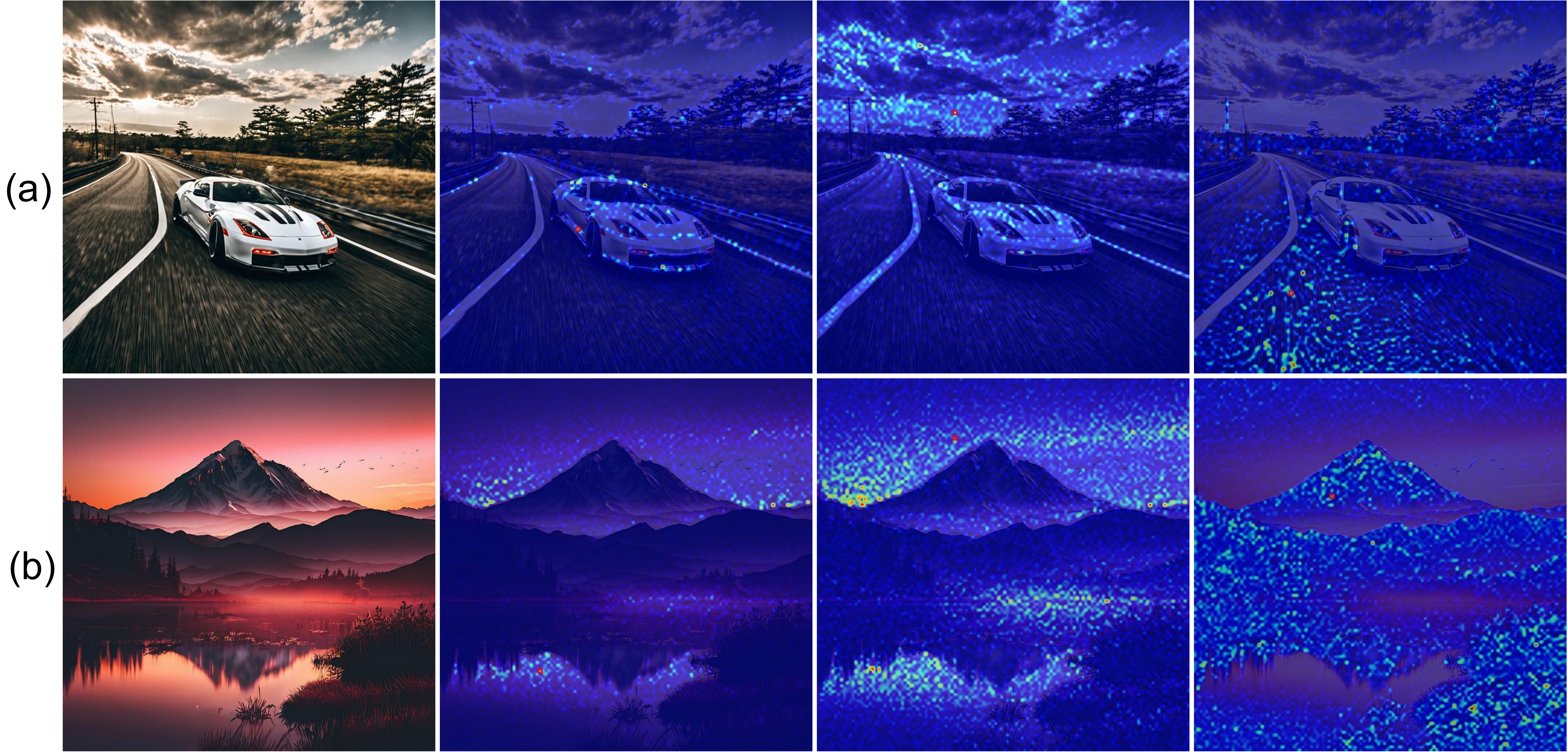}
    \vspace{-2em}
    \caption{\textbf{Visualization of attention maps in TFSA.} The query tokens are highlighted with red boxes, and the heatmap color ranges from blue to red, indicating increasing correlation strength between the key tokens and the query tokens. Resolution: $1024\times 1024$. Zoom-in for a better view.}
    \label{fig:visualize_attn_map}
    \vspace{-1em}
\end{figure}

\subsection{Visualization of Attention Maps in TFSA}
To further demonstrate the clustering effect of TFSA on related tokens, we visualize its attention maps.
As shown in Fig.~\ref{fig:visualize_attn_map}, without using projection matrices, the correlations between tokens are determined jointly by their represented colors and semantics.
For example, in Fig.~\ref{fig:visualize_attn_map}(a), the key tokens correlated with the query token at the selected car location are related not only to the car itself (\ie, the concept of the car) but also to its color.
TFSA leverages such correlations to fuse token information, thereby accelerating the formation of the overall image layout.

\section{Supplementary Qualitative Comparison of \S\ref{sec:qualitative_exp}}
\label{apx:supplementary_qualitative_exp}
\begin{figure*}[!thb]
    \centering
    \includegraphics[width=1\textwidth]{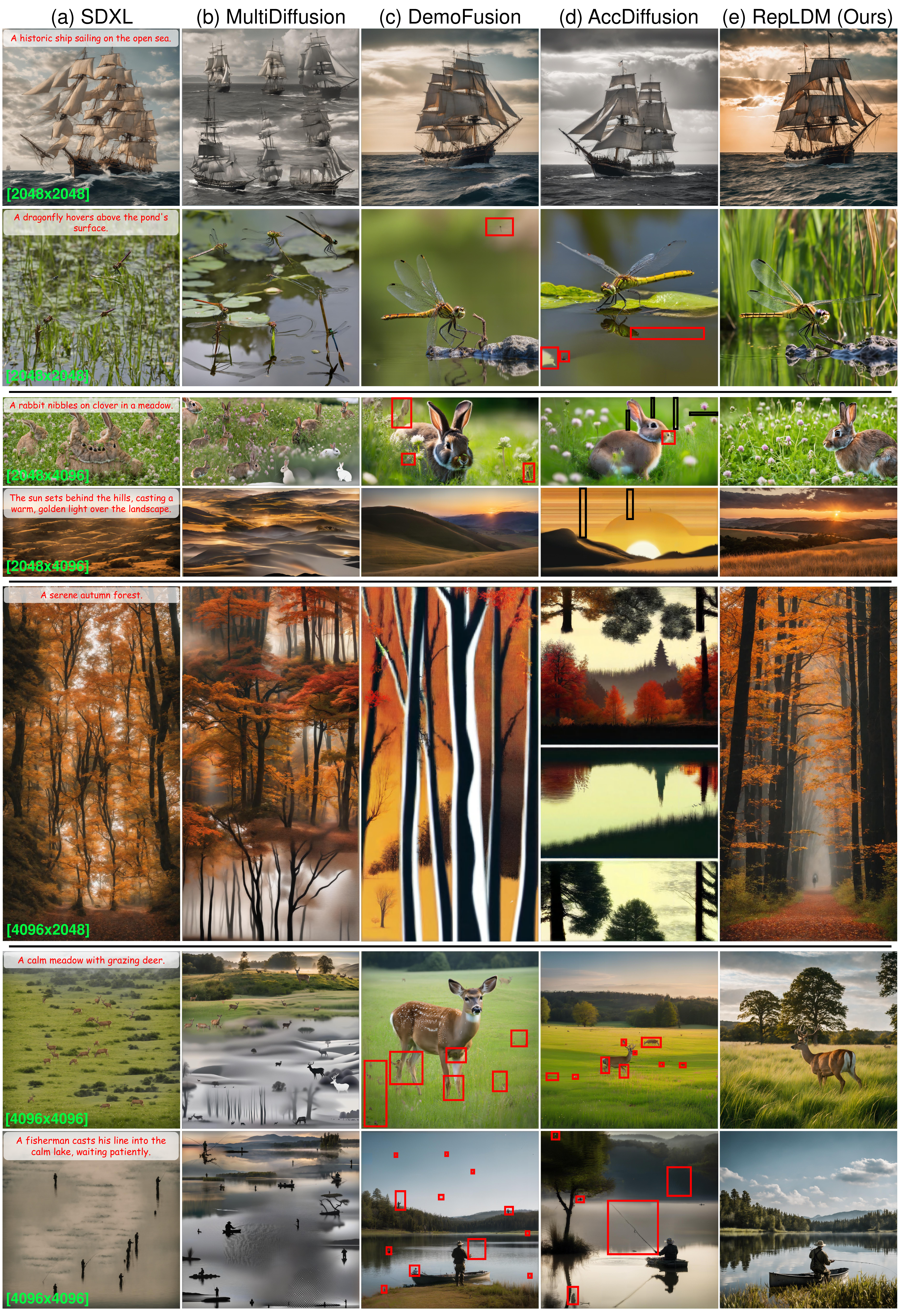}
    \vspace{-2.em}
    \caption{
        \textbf{Qualitative comparison with other baselines.}
        Zoom-in for a better view.
    }
    \label{fig:appendix_comparison}
    \vspace{-2em}
\end{figure*}

Fig.~\ref{fig:appendix_comparison} presents additional qualitative comparison results. MultiDiffusion continues to struggle with maintaining global consistency; as indicated by the red boxes, DemoFusion tends to produce repetitive content, a problem somewhat alleviated in AccDiffusion but not fully resolved. As highlighted by the black boxes, another issue with AccDiffusion is the presence of noticeable streak artifacts in the images.

\section{Supplementary Ablation Experiments of \S\ref{sec:ablation}}
\label{apx:supplementary_ablation}

\subsection{Further Qualitative Analysis of Attention Guidance}
\begin{figure*}[!ht]
\vspace{-.5em}
\centering
\includegraphics[width=1\textwidth]{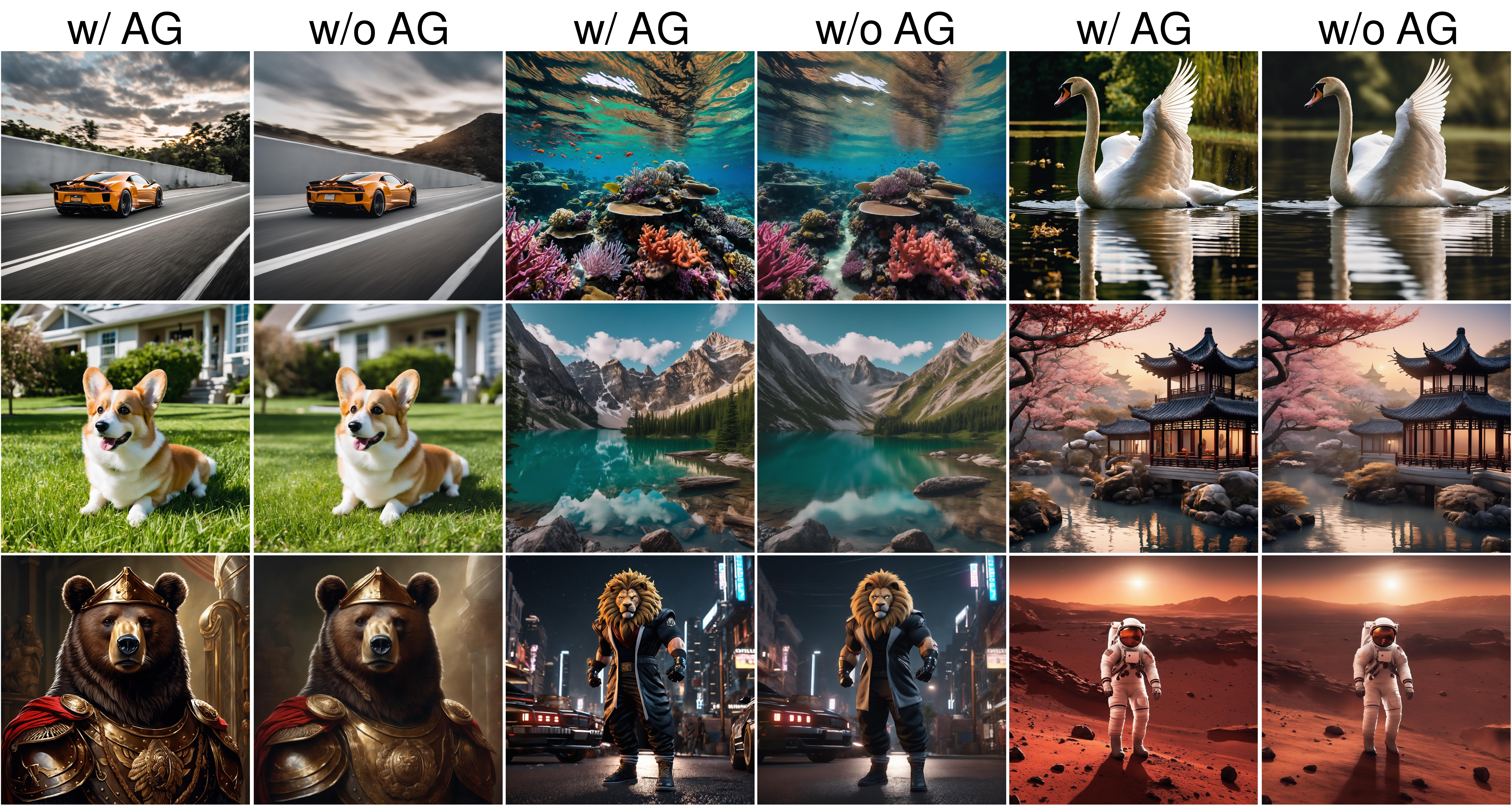}
\vspace{-2em}
\caption{\textbf{Further qualitative analysis of \moduleone (AG).} Using \moduleone significantly enhances image quality. The details were enriched, for example: the clouds in the sky, ripples on the water, reflections on the lake, and even the expressions in a person's eyes. Resolution: 2048$\times$2048. Best viewed \textbf{ZOOMED-IN}.}
\label{fig:appendix_ablation_ag}
\vspace{-.5em}
\end{figure*}

Fig.~\ref{fig:appendix_ablation_ag} provides additional qualitative ablation results on \moduleone. Individual preferences for contrast, color vividness, and detail richness may vary. \moduleone allows users to adjust parameters such as the guidance scale to synthesize images according to their preferences.

\subsection{Ablation on the hyper-parameters of Attention Guidance}
\label{apx:supplementary_ablation_ag}
\paragraph{Quantitative analysis of guidance scale.} We sampled 1k prompts, fixed $\eta_1=0.06, \eta_2=[0.2]$ and performed ablation studies for guidance scale $\gamma$. The quantitative results are shown in Table~\ref{table:ablate_guidance_scale}. Considering all metrics, we find that $\gamma=0.004$ achieved better quantitative results.

\begin{table*}[ht!]
\tiny
\renewcommand{\arraystretch}{1.2}
\centering
\caption{\textbf{Quantitative ablation experiments on the guidance scale $\gamma$}. The best results are marked in \textbf{bold}, and the second best results are marked by \underline{underline}.}
\begin{adjustbox}{width=1\linewidth,center}
\begin{tabular}{r|rrrrr|rrrrr|rrrrr}
\toprule[1pt]
\multirow{2}{*}{\textbf{Method}} & \multicolumn{5}{c|}{$1024 \times 1024$} & \multicolumn{5}{c|}{$1600 \times 1600$} & \multicolumn{5}{c}{$2048 \times 2048$} \cr\cline{2-16}
& FID $\downarrow$ & IS $\uparrow$ & FID$_{c}$ $\downarrow$ & IS$_{c}$ $\uparrow$ & CLIP $\uparrow$
& FID $\downarrow$ & IS $\uparrow$ & FID$_{c}$ $\downarrow$ & IS$_{c}$ $\uparrow$ & CLIP $\uparrow$
& FID $\downarrow$ & IS $\uparrow$ & FID$_{c}$ $\downarrow$ & IS$_{c}$ $\uparrow$ & CLIP $\uparrow$ \\
\midrule
$\gamma=0.000$
& 90.85 & 58.18 & 21.21 & \textbf{17.69} & 25.09
& 90.91 & 54.74 & \textbf{21.45} & 15.41 & 24.93
& 91.78 & 59.08 & \underline{21.57} & \underline{17.36} & 24.86\\
$\gamma=0.001$
& 90.50 & 58.04 & \textbf{21.34} & 16.76 & 25.08
& 91.17 & 54.31 & 21.19 & \underline{15.47} & 24.93
& 91.40 & 58.75 & \textbf{21.87} & 15.85 & 24.86\\
$\gamma=0.002$
& 89.82 & 57.54 & \underline{21.28} & \underline{17.04} & 25.08
& 90.39 & \textbf{53.71} & 21.26 & 15.00 & 24.97
& 90.81 & \underline{58.34} & 21.45 & 17.16 & 24.90\\
$\gamma=0.003$
& 90.10 & \underline{57.08} & 20.80 & 16.61 & 25.08
& 90.56 & \underline{53.95} & \underline{21.35} & 15.46 & 24.98
& 90.87 & 58.40 & 21.47 & \textbf{17.60} & 24.92\\
$\gamma=0.004$
& \textbf{89.40} & \textbf{56.64} & 20.96 & 16.63 & 25.09
& \textbf{89.91} & 54.23 & 20.91 & \textbf{15.54} & 25.01
& \textbf{90.11} & \textbf{58.11} & 21.18 & 16.78 & 24.94s\\
$\gamma=0.005$
& 90.17 & 57.50 & 20.89 & 16.34 & 25.12
& \underline{90.24} & 55.19 & 20.67 & 15.21 & 25.02
& 90.46 & 58.91 & 20.79 & 16.87 & 24.97\\
$\gamma=0.006$
& \underline{89.79} & 58.18 & 20.33 & 15.93 & 25.16
& 90.36 & 56.71 & 20.33 & 14.59 & 25.06
& \underline{90.32} & 59.86 & 20.37 & 16.12 & 25.00\\
$\gamma=0.007$
& 90.42 & 60.29 & 20.07 & 16.20 & 25.21
& 90.91 & 59.35 & 20.36 & 14.16 & 25.12
& 90.86 & 61.81 & 20.14 & 15.70 & 25.06\\
$\gamma=0.008$
& 91.64 & 63.63 & 19.66 & 14.25 & \textbf{25.25}
& 91.98 & 63.93 & 19.13 & 13.71 & \underline{25.13}
& 92.16 & 64.82 & 19.59 & 14.24 & \underline{25.08}\\
$\gamma=0.009$
& 94.29 & 67.87 & 19.15 & 13.00 & \underline{25.25}
& 94.38 & 70.21 & 19.45 & 12.12 & \textbf{25.16}
& 94.39 & 68.84 & 19.22 & 13.63 & \textbf{25.12}\\
\midrule[1pt]
\end{tabular}
\end{adjustbox}
\label{table:ablate_guidance_scale}
\vspace{-1em}
\end{table*}

\paragraph{Quantitative analysis of delay rate.} We sampled 1k prompts, fixed $\gamma=0.004, \eta_2=[0.2]$ and performed ablation studies for delay rate $\eta_1$. Table~\ref{table:ablate_delay_rate} presents the experimental results, indicating that better results can be achieved when $\eta_1 = 0.06$. 
This means that appropriately delaying the effect of \moduleone can further enhance the quality of the generated images.

\begin{table*}[ht!]
\vspace{-.5em}
\tiny
\renewcommand{\arraystretch}{1.2}
\centering
\caption{\textbf{Quantitative ablation experiments on the delay rate $\eta_1$}. The best results are marked in \textbf{bold}, and the second best results are marked by \underline{underline}.}
\begin{adjustbox}{width=1\linewidth,center}
\begin{tabular}{r|rrrrr|rrrrr|rrrrr}
\toprule[1pt]
\multirow{2}{*}{\textbf{Method}} & \multicolumn{5}{c|}{$1024 \times 1024$} & \multicolumn{5}{c|}{$1600 \times 1600$} & \multicolumn{5}{c}{$2048 \times 2048$} \cr\cline{2-16}
& FID $\downarrow$ & IS $\uparrow$ & FID$_{c}$ $\downarrow$ & IS$_{c}$ $\uparrow$ & CLIP $\uparrow$
& FID $\downarrow$ & IS $\uparrow$ & FID$_{c}$ $\downarrow$ & IS$_{c}$ $\uparrow$ & CLIP $\uparrow$
& FID $\downarrow$ & IS $\uparrow$ & FID$_{c}$ $\downarrow$ & IS$_{c}$ $\uparrow$ & CLIP $\uparrow$ \\
\midrule
$\eta_1=0.00$
& 89.98 & 58.29 & 20.74 & 16.48 & 25.06
& 90.89 & 55.54 & 21.00 & 14.42 & 24.98
& 90.75 & 59.41 & 20.54 & 16.99 & 24.91\\
$\eta_1=0.02$
& 89.96 & 57.67 & 20.99 & \textbf{16.87} & 25.05
& 90.76 & 54.77 & 21.08 & 15.35 & 24.95
& 91.78 & 59.08 & \textbf{21.57} & \textbf{18.16} & 24.86\\
$\eta_1=0.04$
& 89.47 & 57.28 & 20.98 & 16.63 & 25.07
& 90.22 & 54.14 & 20.86 & 15.43 & 24.98
& 90.52 & 58.47 & 20.76 & 17.02 & 24.91\\
$\eta_1=0.06$
& \underline{89.44} & \textbf{56.64} & 20.92 & 16.58 & \textbf{25.11}
& \underline{89.91} & 54.23 & 20.91 & 15.54 & 25.01
& \textbf{90.11} & \textbf{58.11} & 21.18 & 16.78 & 24.94\\
$\eta_1=0.08$
& 89.95 & 56.97 & 21.05 & 16.76 & 25.09
& \textbf{89.87} & 54.10 & 21.22 & \underline{15.65} & 24.98
& 90.74 & 58.45 & 20.99 & 17.06 & \textbf{24.92}\\
$\eta_1=0.10$
& \textbf{89.29} & \underline{56.88} & \textbf{21.11} & \underline{16.84} & 25.09
& 89.97 & 53.99 & 21.04 & 15.37 & \underline{24.99}
& 90.41 & 58.45 & 20.99 & 17.12 & \underline{24.92}\\
$\eta_1=0.12$
& 89.84 & 57.32 & 21.05 & 16.58 & 25.08
& 90.00 & 53.85 & 21.24 & \textbf{15.81} & 24.93
& \underline{90.24} & 58.45 & \underline{21.24} & \underline{17.36} & 24.90\\
$\eta_1=0.14$
& 89.85 & 57.12 & 20.91 & 16.40 & \underline{25.09}
& 90.06 & \underline{53.83} & \underline{21.33} & 15.62 & \textbf{24.99}
& 90.69 & \underline{58.25} & 21.17 & 16.74 & 24.91\\
$\eta_1=0.16$
& 90.06 & 57.28 & \underline{21.10} & 16.53 & 25.09
& 90.91 & 54.74 & \textbf{21.45} & 15.41 & 24.93
& 90.76 & 58.37 & 20.97 & 16.87 & 24.91\\
$\eta_1=0.18$
& 90.16 & 57.29 & 20.88 & 15.10 & 25.08
& 90.26 & \textbf{53.79} & 21.06 & 15.07 & 24.97
& 90.78 & 58.33 & 21.05 & 17.21 & 24.90\\
\midrule[1pt]
\end{tabular}
\end{adjustbox}
\label{table:ablate_delay_rate}
\vspace{-1.em}
\end{table*}

\subsection{Ablation on Progressive Scheduler Value}
\label{apx:supplementary_ablation_progressive}
This section presents the results of quantitative ablation analysis on the progressive scheduler $\eta_2$ in the second stage of \framework.
We fixed $\gamma=0, \eta_1=0$, sampled 500 prompts, and generated 1k images to investigate the optimal value of the progressive scheduler.
Table~\ref{table:ablation_init_rate} presents the quantitative results, indicating that using an excessively large progressive scheduler may lead to a decline in image quality.

\begin{table*}[ht!]
\vspace{-.8em}
\tiny
\renewcommand{\arraystretch}{1.2}
\centering
\caption{\textbf{Quantitative ablation study of the progressive scheduler Value}. The best results are marked in \textbf{bold}, and the second best results are marked by \underline{underline}.}
\begin{adjustbox}{width=0.75\linewidth,center}
\begin{tabular}{r|rrrrr|rrrrr}
\toprule[1pt]
\multirow{2}{*}{\textbf{Method}} & \multicolumn{5}{c|}{$1600 \times 1600$} & \multicolumn{5}{c}{$2048 \times 2048$} \cr\cline{2-11} 
& FID $\downarrow$ & IS $\uparrow$ & FID$_{c}$ $\downarrow$ & IS$_{c}$ $\uparrow$ & CLIP $\uparrow$
& FID $\downarrow$ & IS $\uparrow$ & FID$_{c}$ $\downarrow$ & IS$_{c}$ $\uparrow$ & CLIP $\uparrow$ \\
\midrule
SDXL
& 101.56 & 25.78 & 73.67 & 21.23 & 26.87
& 112.64 & 18.44 & 79.03 & 20.61 & 26.55 \\
$\eta_2=[0.9]$
& 94.59 & 27.04 & 67.60 & 23.01 & 26.97
& 97.14 & 24.48 & 64.34 & 22.14 & 26.59 \\
$\eta_2=[0.8]$
& 93.13 & 28.80 & 65.67 & 24.83 & 26.99
& 93.93 & 26.75 & 60.84 & 23.27 & 26.77 \\
$\eta_2=[0.7]$
& \textbf{92.05} & 29.44 & 65.35 & 24.97 & 27.07
& 92.50 & 28.17 & 57.34 & 24.05 & 26.93 \\
$\eta_2=[0.6]$
& 92.94 & 30.79 & 64.57 & 24.29 & 27.11
& 91.86 & 30.45 & 55.38 & 24.96 & 26.98 \\
$\eta_2=[0.5]$
& \underline{92.73} & 30.65 & 63.43 & 24.26 & 27.13
& \underline{91.80} & \underline{31.18} & 54.32 & 24.48 & 27.02 \\
$\eta_2=[0.4]$
& 93.04 & \underline{30.96} & 63.33 & 24.77 & 27.14
& \textbf{91.71} & \textbf{32.47} & 53.72 & 25.16 & 27.03 \\
$\eta_2=[0.3]$
& 92.93 & 30.91 & \textbf{63.09} & 24.84 & 27.15
& 92.39 & 30.72 & \underline{53.32} & \textbf{26.63} & 27.07 \\
$\eta_2=[0.2]$
& 93.09 & \textbf{31.17} & \underline{63.23} & \textbf{25.71} & \underline{27.17}
& 92.71 & 30.45 & \textbf{53.19} & \underline{26.19} & \underline{27.12} \\
$\eta_2=[0.1]$
& 93.44 & 30.69 & 63.75 & \underline{25.18} & \textbf{27.22}
& 92.94 & 30.69 & 53.77 & 24.71 & \textbf{27.18} \\
\midrule[1pt]
\end{tabular}
\end{adjustbox}
\label{table:ablation_init_rate}
\vspace{-1.75em}
\end{table*}

\section{Ablation on the Attention Guidance Components}

\subsection{Ablation on the Guidance Scale Decay Strategy}
\begin{figure}[!ht]
    \centering
    \includegraphics[width=1\linewidth]{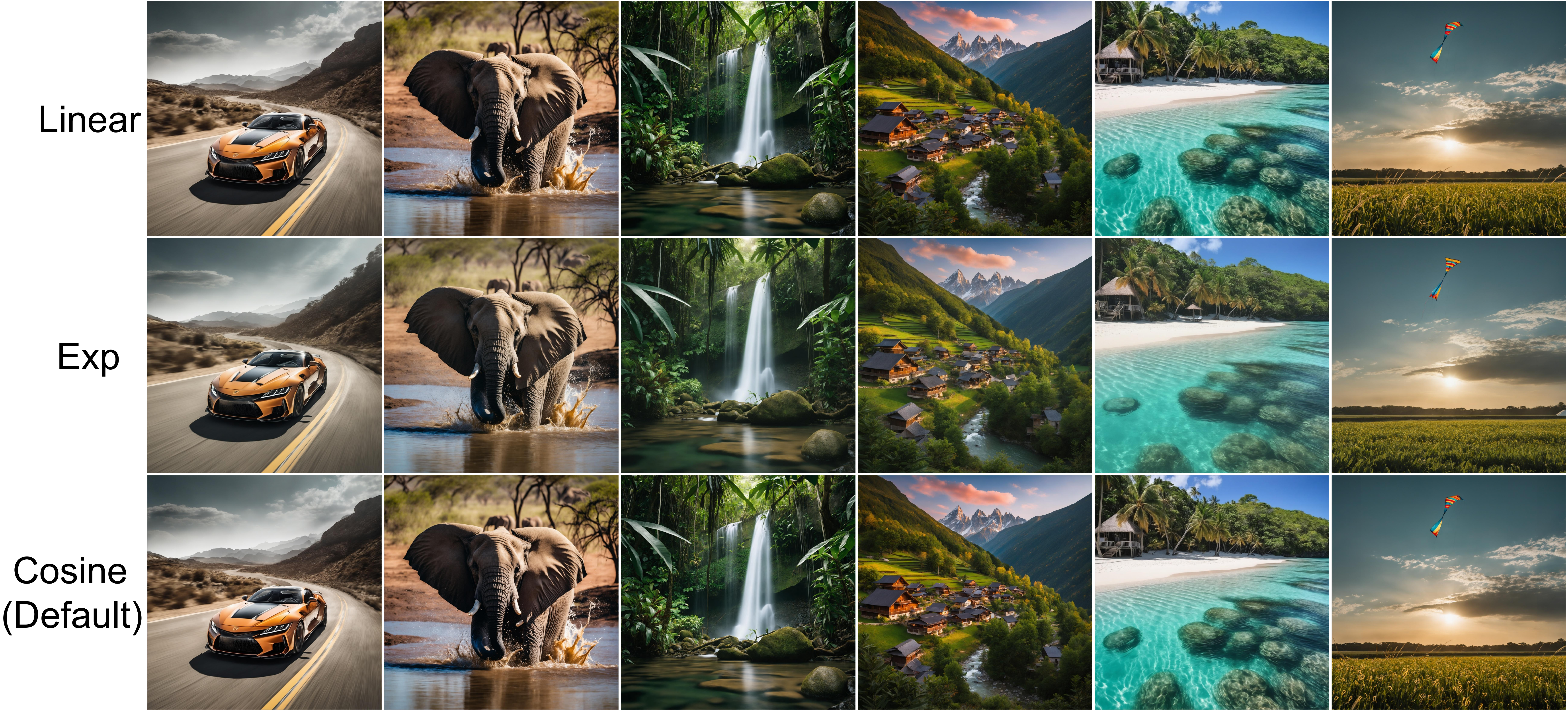}
    \vspace{-2em}
    \caption{\textbf{Qualitative ablation on guidance scale decay ctrategies.}}
    \label{fig:ablate_decay_strategy}
    \vspace{-1em}
\end{figure}

To investigate the impact of different guidance scale decay strategies, we conduct ablation studies using two additional schemes—linear decay and exponential decay—and analyze their quantitative and qualitative performance.
For quantitative ablation, we generate 2k samples at a resolution of $2048\times 2048$ using each strategy and calculate the criterions on the SAM benchmark.
Table~\ref{table:ablate_decay_strategy} shows that different strategies yield similar results, indicating that \framework is not sensitive to a specific decay strategy.
Fig.~\ref{fig:ablate_decay_strategy} illustrates the qualitative results.
Qualitatively, these decay strategies also produce similar visual experience.
\begin{table}[!ht]
\caption{\textbf{Ablation on the guidance scale decay strategies.} The best results are marked in \textbf{bold}, and the second best results are marked by \underline{underline}.}
\vspace{-.5em}
\label{table:ablate_decay_strategy}
\centering
\tiny
\begin{tabular}{rrrrrr}
\toprule[1pt]
\textbf{Strategies} & FID $\downarrow$ & IS$_c$ $\uparrow$ & FID$_c$ $\downarrow$ & IS$_c$ $\uparrow$ & CLIP $\uparrow$\\
\midrule
Linear
& \underline{66.2} & \underline{21.5} & \underline{47.2} & \textbf{20.3} & \textbf{25.4}\\
Exponential
& 66.8 & \textbf{21.8} & \textbf{47.0} & 16.3 & \underline{25.3}\\
Cosine (default)
& \textbf{66.0} & 21.0 & 47.4 & \underline{17.5} & 25.1\\
\bottomrule[1pt]
\end{tabular}
\vspace{-1.25em}
\end{table}

\subsection{Ablation on the Attention Calculation Paradigm}
For TFSA, our objective is to remove the learnable parameters from the Self-Attention mechanism, while maintaining its computational paradigm as unchanged as possible.
In TFSA, $\vct{Q}$, $\vct{K}$, and $\vct{V}$ are identical.
Therefore, TFSA is a totally symmetric formula.
As analyzed before, this paradigm encourages the clustering of semantically related tokens, and finally leads to finer details and richer colors.
An interesting question arises: if we spatially downsample $\vct{Q}$, $\vct{K}$, or $\vct{V}$ before applying TFSA and reformulate it into an asymmetric paradigm (denoted as TFSA-A), would TFSA-A encourage the model to attend more explicitly from fine details to coarse structures?

\begin{figure}[!ht]
    \centering
    \includegraphics[width=1\linewidth]{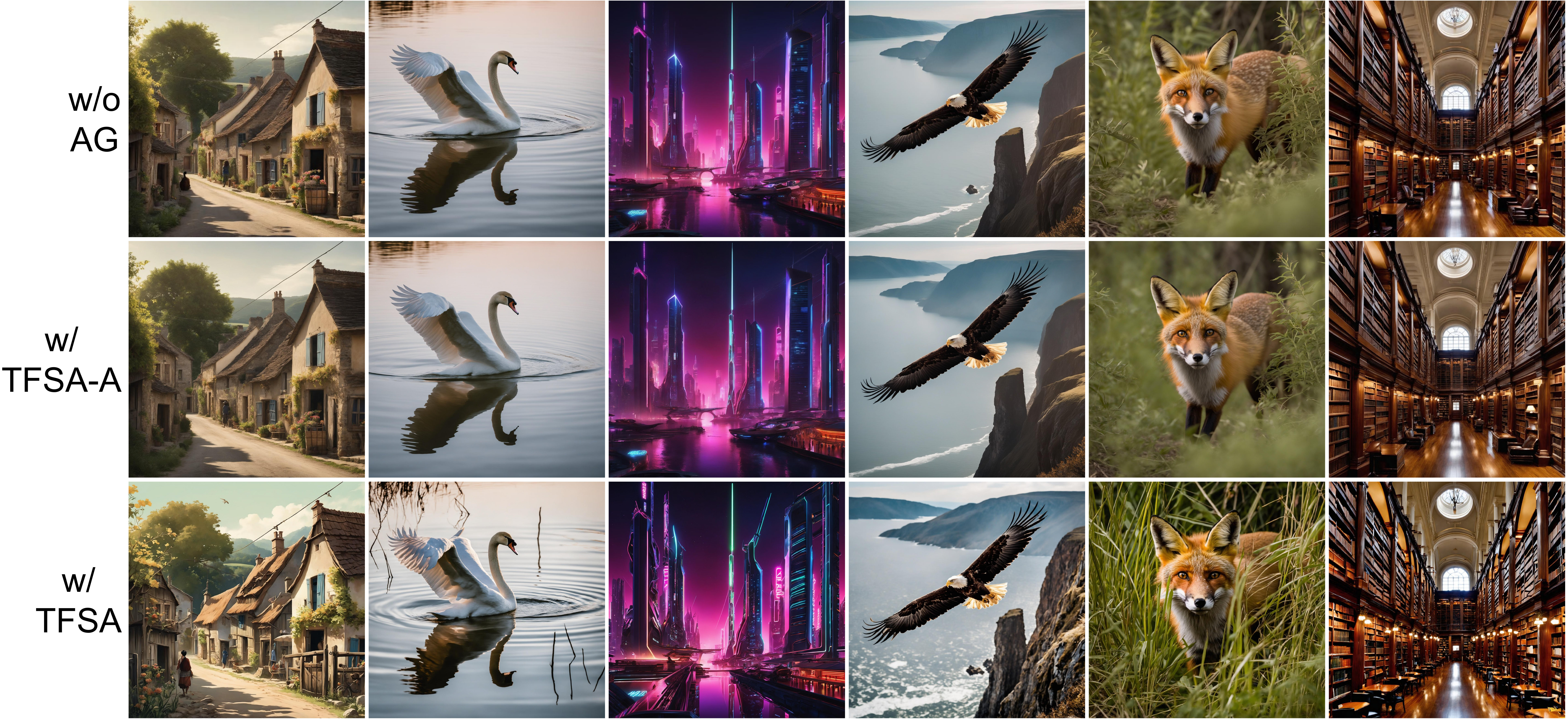}
    \vspace{-2em}
    \caption{\textbf{Ablation on the attention calculation paradigm.} Resolution: $2048\times 2048$.}
    \label{fig:ablate_attention_paradigm}
    \vspace{-1.5em}
\end{figure}

To answer this question, we design an asymmetric variants, TFSA-A.
Specifically, TFSA-A performs a 2$\times$2 pooling operation to downsample the $\vct{K}$ and $\vct{V}$ matrices before the attention calculation operation, ensuring that the output of $\operatorname{Softmax}(\vct{Q}\vct{K}^T/\sqrt{d})\vct{V}$ remains the of shape $(hw) \times c$.
Table~\ref{table:ablate_attention_paradigm} shows that TFSA-A produces comparable quantitative results.
In Fig.~\ref{fig:ablate_attention_paradigm}, we observe that although TFSA-A achieves quantitative results comparable to those of TFSA, its visual quality is significantly inferior.
In fact, TFSA-A tends to reduce image details.
This aligns with our hypothesis: the $2\times 2$ pooling acts as a low-pass filter, causing the loss of fine-grained information in the latent representations and leading the model to focus more on low-frequency structures.
\begin{table}[!ht]
\caption{\textbf{Ablation on the attention calculation paradigm.} The best results are marked in \textbf{bold}, and the second best results are marked by \underline{underline}.}
\vspace{-.5em}
\label{table:ablate_attention_paradigm}
\centering
\tiny
\begin{tabular}{rrrrrr}
\toprule[1pt]
\textbf{Paradigm} & FID $\downarrow$ & IS$_c$ $\uparrow$ & FID$_c$ $\downarrow$ & IS$_c$ $\uparrow$ & CLIP $\uparrow$\\
\midrule
w/o guidance
& \underline{66.8} & \underline{21.6} & \underline{47.5} & 17.4 & \textbf{25.3}\\
w/ TFSA-A
& 67.4 & \textbf{22.6} & 47.9 & \textbf{20.4} & \textbf{25.3}\\
w/ TFSA
& \textbf{66.0} & 21.0 & \textbf{47.4} & \underline{17.5} & \underline{25.1}\\
\bottomrule[1pt]
\end{tabular}
\vspace{-.25em}
\end{table}

\section{Further Model Efficiency Analysis}
\label{apx:apldm_process_analyze}

\paragraph{Computational complexity analysis of TFSA.}
Note that \moduleone is only applied during the first stage of generation.
Assume we have a HR image $\vct{x}_0$ with a resolution of $H \times W \times C$. we encode the image $\vct{x}_0$ into latent space and obtain latent representation $\vct{z}_0 \in \mathbb{R}^{h \times w \times c}$.
Before feeding $\vct{z}_0$ into TFSA, we reshape it to a $(hw) \times c$ matrice.
The computation of TFSA follows a formulation similar to that of self-attention:
$\operatorname{Softmax}(\vct{z}_0\vct{z}_0^T/\sqrt{c})\vct{z}$.
Thus, the computational complexity of TFSA is $O((hw)^2c)$.
Taking SDXL as an example, the training resolution is $H=1024$, $W=1024$.
After VAE encoding, $c=4$, $h=H/8=128$, $w=W/8=128$.
For each denoising step, the FLOPs of TFSA is approximately $2 \times (h \times w)^2 \times c$, which is around 2.15 GFLOPs---negligible compared to the FLOPs of the denoising network (several TFLOPs per step).

\paragraph{How does pixel space upsampling accelerate generation?}
To answer this question, we analyze the time consumption of each component in DemoFusion and \framework when generating images at the resolution of $4096\times 4096$.

\begin{table}[!ht]
\vspace{-.2em}
\caption{\textbf{The time consumption of DemoFusion when generating $4096\times 4096$ resolution images.}}
\vspace{-.5em}
\label{table:encode_decode_cost_demofusion}
\centering
\tiny
\begin{tabular}{rrrrrrr}
\toprule[1pt]
\textbf{Metric} & Denoise 1024 & Denoise 2048 & Denoise 3072 & Denoise 4096 & Decode 4096 & Total\\
\midrule
number of steps
& 50 & 50 & 50 & 50 & - & 200\\
Time (s)
& 12 & 185 & 480 & 901 & 106 & 1684\\
\bottomrule[1pt]
\end{tabular}
\vspace{-1.2em}
\end{table}

\begin{table}[!ht]
\vspace{-.5em}
\caption{\textbf{The time consumption of \framework when generating $4096\times 4096$ resolution images.} The intermediate encoding/decoding operations are highlighted in $\underline{\text{underline}}$.}
\vspace{-.5em}
\label{table:encode_decode_cost_framework}
\centering
\tiny
\begin{adjustbox}{width=1\linewidth,center}
\begin{tabular}{rrrrrrrrrr}
\toprule[1pt]
\textbf{Metric} & Denoise 1024 & Decode 1024 & Encode 3304 & Denoise 3304 & Decode 3304 & Encode 4096 & Denoise 4096 & Decode 4096 & Total\\
\midrule
number of steps
& 50 & - & - & 5 & - & - & 10 & - & 65\\
Time (s)
& 12 & \underline{0} & \underline{12} & 20 & \underline{64} & \underline{11} & 118 & 106 & 343\\
\bottomrule[1pt]
\end{tabular}
\end{adjustbox}
\vspace{-1.5em}
\end{table}

Table~\ref{table:encode_decode_cost_demofusion} shows that denoising at high resolutions is a time-consuming process.
DemoFusion requires substantial generation time because it performs the full denoising process at high resolutions.
Note that, compared with the cost of the denoising process at high resolutions, the costs of encoding and decoding are negligible.
Table~\ref{table:encode_decode_cost_framework} shows that \framework significantly accelerates generation by substantially reducing the number of denoising steps at high resolutions.
This is because \framework performs pixel space upsampling through multiple rounds of encoding and decoding, producing high-quality low-resolution images that serve as better initialization.
As a result, \framework can significantly reduce the number of sampling steps required for HR generation, thereby accelerating the process.
Moreover, Table~\ref{table:encode_decode_cost_framework} shows that the additional overhead from multiple intermediate encoding and decoding operations is also relatively minor compared to the total generation cost.

\paragraph{Further efficiency comparison across different models.}
To provide a more comprehensive assessment of model efficiency, we further report the NFE and FLOPs of different models when generating a single image at resolutions of $2048\times2048$ and $4096\times4096$.
Tables~\ref{table:NFE_FLOPs_2048} and \ref{table:NFE_FLOPs_4096} show that \framework significantly reduces the NFE and FLOPs required for inference by decreasing the number of denoising steps at high resolutions, thereby substantially reducing the time needed to generate HR images.

\begin{table}[!ht]
\vspace{-.5em}
\caption{\textbf{Inference cost of generating a $2048\times 2048$ Image for different models.}}
\vspace{-.5em}
\label{table:NFE_FLOPs_2048}
\centering
\tiny
\begin{adjustbox}{width=1\linewidth,center}
\begin{tabular}{rrrrrrrrr}
\toprule[1pt]
\textbf{Model} & SDXL~\cite{sdxl} & MultiDiff.~\cite{bar2023multidiffusion} & ScaleCrafter~\cite{he2023scalecrafter}& HiDiff.~\cite{zhang2025hidiffusion} & UG~\cite{hwang2024upsample} & DemoFusion~\cite{du2024demofusion} & AccDiff.~\cite{lin2024accdiffusion} & \framework\\
\midrule
NFE
& 50 & 50 & 50 & 50 & 80 & 100 & 100 & 60\\
TFLOPs
& 3010 & 5420 & 2437 & 1857 & 3608 & 9015 & 8597 & 1140\\
Time (min)
& 1.0 & 3.0 & 1.0 & 0.8 & 1.8 & 3.0 & 3.0 & 0.6\\
\bottomrule[1pt]
\end{tabular}
\end{adjustbox}
\vspace{-1.5em}
\end{table}

\begin{table}[!ht]
\caption{\textbf{Inference cost of generating a $4096\times 4096$ Image for different models.}}
\vspace{-.5em}
\label{table:NFE_FLOPs_4096}
\centering
\tiny
\begin{adjustbox}{width=1\linewidth,center}
\begin{tabular}{rrrrrrrrr}
\toprule[1pt]
\textbf{Model} & SDXL~\cite{sdxl} & MultiDiff.~\cite{bar2023multidiffusion} & ScaleCrafter~\cite{he2023scalecrafter}& HiDiff.~\cite{zhang2025hidiffusion} & UG~\cite{hwang2024upsample} & DemoFusion~\cite{du2024demofusion} & AccDiff.~\cite{lin2024accdiffusion} & \framework\\
\midrule
NFE
& 50 & 50 & 50 & 50 & 80 & 200 & 200 & 65\\
TFLOPs
& 12026 & 29566 & 9759 & 5211 & 12624 & 72167 & 74225 & 7140\\
Time (min)
& 8.0 & 15.0 & 19.0 & 3.4 & 11.1 & 25.0 & 26.0 & 5.7\\
\bottomrule[1pt]
\end{tabular}
\end{adjustbox}
\vspace{-1.5em}
\end{table}

\begin{figure*}[!th]
\vspace{-.5em}
\centering
\includegraphics[width=1\textwidth]{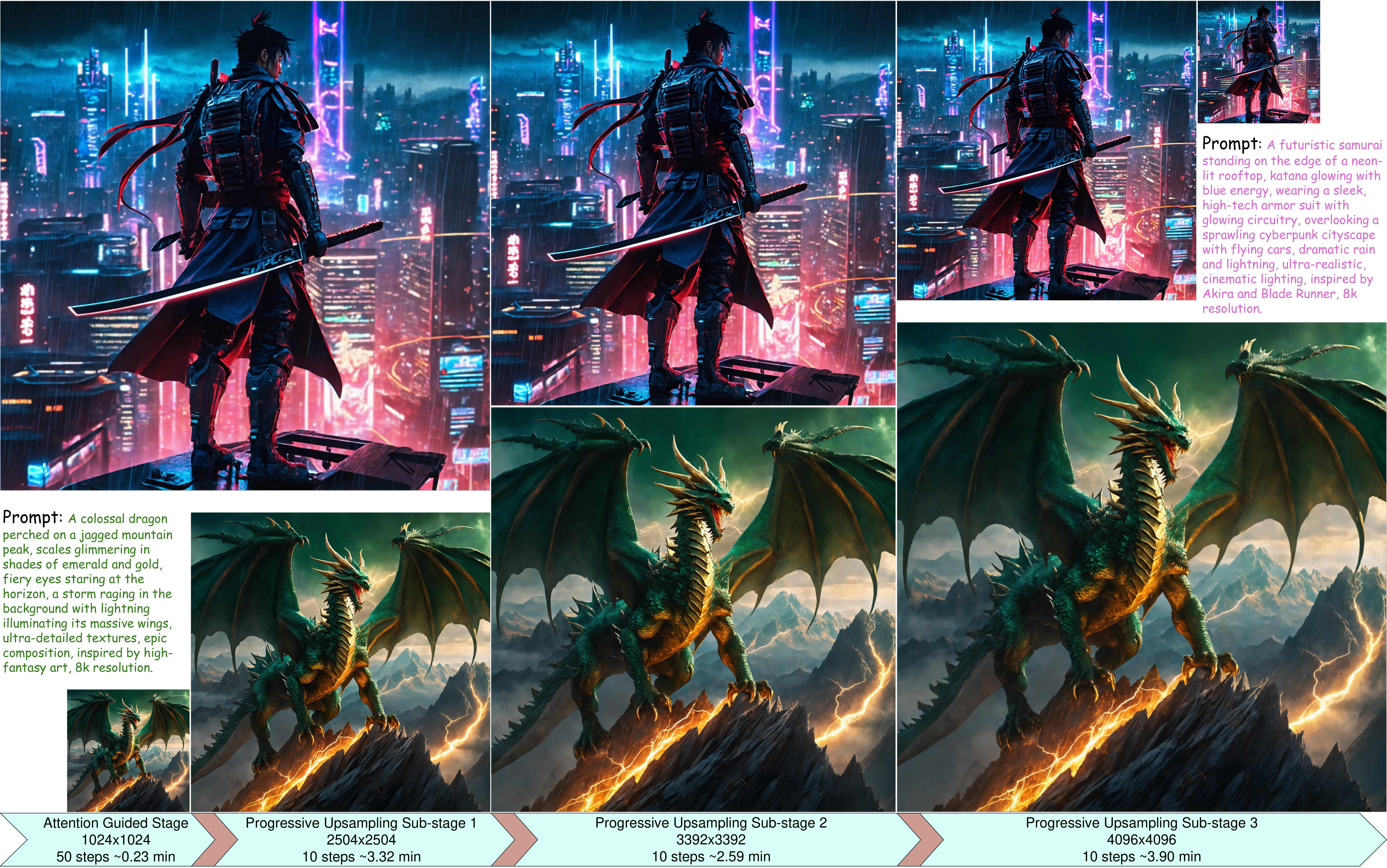}
\vspace{-2em}
\caption{\textbf{Illustration of the progressive upsampling generation process.} The inference speed is evaluated on a single 3090 GPU.}
\vspace{-1.em}
\label{fig:appendix_progressive}
\end{figure*}

\paragraph{Qualitative analysis on the progressive upsampling stage.}
To clearly illustrate the progressive upsampling process of \framework, we set \(\eta_2 = [0.2, 0.2, 0.2]\) to generate \(4096 \times 4096\) images. As shown in Fig.~\ref{fig:appendix_progressive}, the images generated at different sub-stages of \framework exhibit a high degree of consistency, with only minor differences in details. Since our task focuses on generating HR images rather than traditional image super-resolution, these differences in details are reasonable.
As discussed in Table~\ref{table:encode_decode_cost_demofusion} and Table~\ref{table:encode_decode_cost_framework}, for each denoising step, the time required for HR images is several times that for low-resolution images. Consequently, repeating a full denoising process at high resolution is extremely time-consuming~\citep{du2024demofusion, lin2024accdiffusion}.
Considering that HR and low-resolution images should share the same low-frequency structure, and that DMs naturally generate low-frequency structures first during denoising~\citep{yu2023freedom, teng2023relay}, \framework leverages the prior knowledge of low-frequency structures in low-resolution images, thereby effectively accelerating the generation process.

\section{\framework Algorithm}
\label{apx:framework_algorithm}
The implementation details of \framework can be found in Algorithm~\ref{al:swf_pipeline}, and further information is available in our code repository.

\begin{algorithm*}[!ht]
\footnotesize
\caption{\framework Inference Pipeline}
\label{al:swf_pipeline}
\begin{algorithmic}[1]
\REQUIRE The number of inference time steps of the first stage $T_0$; progressive scheduler $\eta_2$; \moduleone scale $\gamma$; \moduleone delay rate $\eta_1$; the decay factor $\beta$; target image size tuple $(H^{\prime}, W^{\prime})$; the denoising model $\boldsymbol{\mathcal{F}}$; denoising model's training resolution tuple $(H, W)$; VAE encoder $\boldsymbol{\mathcal{E}}$; VAE decoder $\boldsymbol{\mathcal{D}}$; noise scheduler's hyper-parameter list $\bar{\alpha}_{1:T_0}$.
\STATE \textbf{Initialization:}
\STATE $\boldsymbol{z}_{T_0}^{(0)} = \boldsymbol{\epsilon} \sim \mathcal{N}(0, \boldsymbol{I})$ \COMMENT{Sampling from Standard Gaussian Distribution}
\STATE $n_\text{stages} = \text{length}(\eta_2) + 1$ \COMMENT{Get the total number of denoising stages}
\STATE $r^{\prime} = \frac{H^{\prime}}{W^{\prime}}$ \COMMENT{Keep the aspect ratio and number of pixels unchanged}
\STATE $H^{(0)} = \text{ceil}(\sqrt{H\times W\times r^{\prime}})$
\STATE $W^{(0)} = \text{ceil}(\sqrt{\frac{H\times W}{r^{\prime}}})$
\STATE $H^{(n)} = H^\prime$
\STATE $W^{(n)} = W^\prime$
\STATE $area_\text{list} = \text{linspace}(H^{(0)}\times W^{(0)}, H^{(n)}\times W^{(n)}, n_\text{stages})$ \COMMENT{Upsampling according to the number of pixels}
\STATE $H_\text{list} = [\text{ceil}(\sqrt{i\times r^\prime})\quad \text{for}\quad i\quad \text{in}\quad area_\text{list}]$ \COMMENT{Get the height and width of each stage}
\STATE $W_\text{list} = [\text{ceil}(\sqrt{i/r^\prime})\quad \text{for}\quad i\quad \text{in}\quad area_\text{list}]$
\STATE $k_\text{denoising}=[T_0]$ \COMMENT{Get the number of denoising steps for each stage}
\STATE $k_\text{denoising}.\text{extend}([i\times T_0\quad \text{for}\quad i\quad \text{in}\quad \eta_2])$
\STATE $k = T_0\times \eta_1$ \COMMENT{Obtain the number of delay steps}
\STATE $\gamma_\text{list} = [\gamma(\frac{cos(\frac{T-k-i}{T-k}\pi)+1}{2})^\beta\quad \text{for}\quad i=1,...,T-k]$ \COMMENT{Obtain the guidance scale for each step}
\STATE \textbf{Denoising:}
\FOR{$s = 0,\ldots,n_\text{stages}-1$}
    \STATE $n_\text{steps} \gets k_\text{denoising}[s]$
    \IF{$s \geq 1$}
        \STATE $\boldsymbol{x}^{(s)} \gets \text{upsample}(\boldsymbol{x}^{(s-1)}, H_\text{list}[s], W_\text{list}[s])$ \COMMENT{Upsampling in pixel space}
        \STATE $\boldsymbol{z}_0^{(s)} \gets \boldsymbol{\mathcal{E}}(\boldsymbol{x}^{(s)})$
        \STATE $\boldsymbol{z}_{n_\text{steps}}^{(s)} \sim \mathcal{N}(\sqrt{\bar{\alpha}[n_\text{steps}]}\boldsymbol{z}_0^{(s)}, (1-\bar{\alpha}[n_\text{steps}])\boldsymbol{I})$
    \ENDIF
    \FOR{$t = n_\text{steps}-1,\ldots,0$}
        \STATE $\boldsymbol{z}_t^{(s)} \gets \boldsymbol{\mathcal{F}}(\boldsymbol{z}_{t+1}^{(s)}, t+1)$ \COMMENT{Denoising}
        \IF{$s == 0$ and $t \le T - 1 - k$}
            \STATE $\boldsymbol{z}_t^{(s)} \gets \gamma_\text{list}[t] \text{PFSA}(\boldsymbol{z}_t^{(s)}) + (1 - \gamma_\text{list}[t]) \boldsymbol{z}_t^{(s)}$ \COMMENT{Attention Guidance}
        \ENDIF
    \ENDFOR
    \STATE $\boldsymbol{x}^{(s)} \gets \boldsymbol{\mathcal{D}}(\boldsymbol{z}_0^{(s)})$ \COMMENT{Obtain the pixel space image}
\ENDFOR
\end{algorithmic}
\end{algorithm*}

\section{Robustness Analysis}
\label{apx:robustness_analysis}
In this section, we conduct a robustness analysis to complement the experiments in \S\ref{sec:quantitative_exp}, providing a more comprehensive evaluation of the models' performance.
Our robustness analysis is conducted from two perspectives:
\first we vary the random seeds and repeat each experiment three times to compute the mean and standard deviation of all results;
\second we randomly sample 20k HR images from the HR subset of LAION-5B dataset~\cite{schuhmann2022laion} to construct a new benchmark for evaluating the models’ generalization performance.
Since HR generation requires substantial computational resources, we analyze the four best-performing models from Table~\ref{table:comparison}, \ie, HiDiffusion, DemoFusion, AccDiffusion, and \framework.

\paragraph{Analysis on the SAM benchmark.}
We maintain the exact experimental settings as in \S\ref{sec:quantitative_exp} and conduct the analysis at resolutions of $2048\times 2048$ and $4096\times 4096$.
Table~\ref{table:robust_analysis_sam} shows that \framework continues to exhibit superior performance across the repeated experiments.

\begin{table*}[ht!]
\small
\renewcommand{\arraystretch}{1.2}
\centering
\caption{\textbf{Robustness analysis on the SAM benchmark}. The best results are marked in \textbf{bold}.}
\begin{adjustbox}{width=1\linewidth,center}
\begin{tabular}{rrrrrrrrrrr}
\toprule[1.5pt]
\multirow{2}{*}{\textbf{Method}} & \multicolumn{5}{c|}{$2048\times 2048$} & \multicolumn{5}{c}{$4096\times 4096$} \cr\cline{2-11} 
& FID $\downarrow$ & IS $\uparrow$ & FID$_{c}$ $\downarrow$ & IS$_{c}$ $\uparrow$ & CLIP $\uparrow$
& FID $\downarrow$ & IS $\uparrow$ & FID$_{c}$ $\downarrow$ & IS$_{c}$ $\uparrow$ & CLIP $\uparrow$ \\
\midrule[1pt]
HiDiff.~\cite{zhang2025hidiffusion}
& 80.29\pn0.57 & 17.18\pn0.40 & 63.55\pn0.63 & 15.26\pn0.76 & 24.95\pn0.04
& 144.24\pn0.84 & 12.71\pn0.14 & 146.62\pn0.32 & 7.48\pn0.28 & 21.18\pn0.05\\
DemoF.~\cite{du2024demofusion}
& 71.89\pn0.60 & 22.10\pn0.37 & 53.58\pn0.22 & 19.21\pn0.27 & 25.21\pn0.01
& 101.83\pn0.49 & 20.81\pn0.11 & 63.60\pn0.46 & 14.92\pn1.24 & \textbf{24.75}\pn0.03\\
AccDiff.~\cite{lin2024accdiffusion}
& 71.37\pn0.48 & 21.21\pn0.32 & 53.04\pn0.33 & 19.24\pn1.72 & 25.13\pn0.01
& 102.41\pn1.40 & 19.88\pn0.24 & 65.86\pn0.17 & 12.73\pn0.71 & 24.65\pn0.02\\
\framework
& \textbf{66.08}\pn0.02 & \textbf{22.13}\pn0.74 & \textbf{47.31}\pn0.11 & \textbf{20.38}\pn2.03 & \textbf{25.30}\pn0.12
& \textbf{91.46}\pn0.61 & \textbf{21.63}\pn0.46 & \textbf{58.93}\pn0.20 & \textbf{15.02}\pn0.16 & 24.62\pn0.02\\
\bottomrule[1.5pt]
\end{tabular}
\end{adjustbox}
\label{table:robust_analysis_sam}
\vspace{-.5em}
\end{table*}

\paragraph{Analysis on the LAION-5B benchmark.}
Considering that only 1K samples were used for the $4096\times 4096$ resolution in \S\ref{sec:quantitative_exp}, which may lead to unstable metric evaluations, we double the number of samples to 2k for this resolution in the current experiment.
Regarding evaluation metrics, since IS may lead to high variances beyond ImageNet, we follow some recent studies and adopt Kernel Inception distance (KID) for more accurate evaluation~\cite{huang2024fouriscale, qiu2024freescale}.
Table~\ref{table:robust_analysis_laion} shows that on the LAION benchmark, \framework still demonstrates superior performance, surpassing previous SOTA models across all metrics.

\begin{table*}[ht!]
\vspace{-.75em}
\small
\renewcommand{\arraystretch}{1.2}
\centering
\caption{\textbf{Robustness analysis on the LAION-5B benchmark}. The best results are marked in \textbf{bold}. Since the magnitude of KID is relatively small, we multiply its mean and standard deviation by $10^3$.}
\begin{adjustbox}{width=1\linewidth,center}
\begin{tabular}{rrrrrrrrrrr}
\toprule[1.5pt]
\multirow{2}{*}{\textbf{Method}} & \multicolumn{5}{c|}{$2048\times 2048$} & \multicolumn{5}{c}{$4096\times 4096$} \cr\cline{2-11} 
& FID $\downarrow$ & KID$\downarrow$ & FID$_{c}$ $\downarrow$ & KID$_{c}$ $\downarrow$ & CLIP $\uparrow$
& FID $\downarrow$ & KID$\downarrow$ & FID$_{c}$ $\downarrow$ & KID$_{c}$ $\downarrow$ & CLIP $\uparrow$ \\
\midrule[1pt]
HiDiff.~\cite{zhang2025hidiffusion}
& 48.17\pn0.41 & 8.06\pn0.20 & 36.26\pn0.37 & 10.93\pn0.11 & 23.16\pn0.03
& 92.81\pn0.78 & 35.36\pn0.60 & 120.26\pn0.91 & 103.45\pn0.27 & 18.55\pn0.06\\
DemoF.~\cite{du2024demofusion}
& 34.15\pn0.31 & 4.50\pn0.05 & 21.38\pn0.17 & 6.80\pn0.06 & 25.44\pn0.02
& 37.03\pn0.27 & 5.71\pn0.14 & 30.77\pn0.36 & 16.12\pn0.22 & 25.12\pn0.04\\
AccDiff.~\cite{lin2024accdiffusion}
& 34.49\pn0.31 & 4.92\pn0.08 & 22.71\pn0.17 & 8.57\pn0.11 & 24.90\pn0.02
& 38.56\pn0.23 & 7.21\pn0.20 & 38.85\pn0.29 & 20.87\pn0.20 & 24.46\pn0.01\\
\framework
& \textbf{34.08}\pn0.25 & \textbf{4.18}\pn0.04 & \textbf{20.30}\pn0.30 & \textbf{4.87}\pn0.13 & \textbf{25.78}\pn0.03
& \textbf{34.01}\pn0.26 & \textbf{4.13}\pn0.05 & \textbf{23.08}\pn0.26 & \textbf{12.08}\pn0.13 & \textbf{25.88}\pn0.04\\
\bottomrule[1.5pt]
\end{tabular}
\end{adjustbox}
\label{table:robust_analysis_laion}
\vspace{-1.em}
\end{table*}

\section{Comparative and Ablation Analysis Based on StableDiffusion 2.1}
\label{apx:exp_with_sd2.1}

\subsection{Comparison Experiments}

To validate the generalization capability of \framework, we conducted extensive quantitative and qualitative analyses using StableDiffusion 2.1 (SD2.1) as the pretrained base model.

\begin{wrapfigure}{r}{7cm}
\vspace{-1.5em}
\centering
\includegraphics[width=0.48\textwidth]{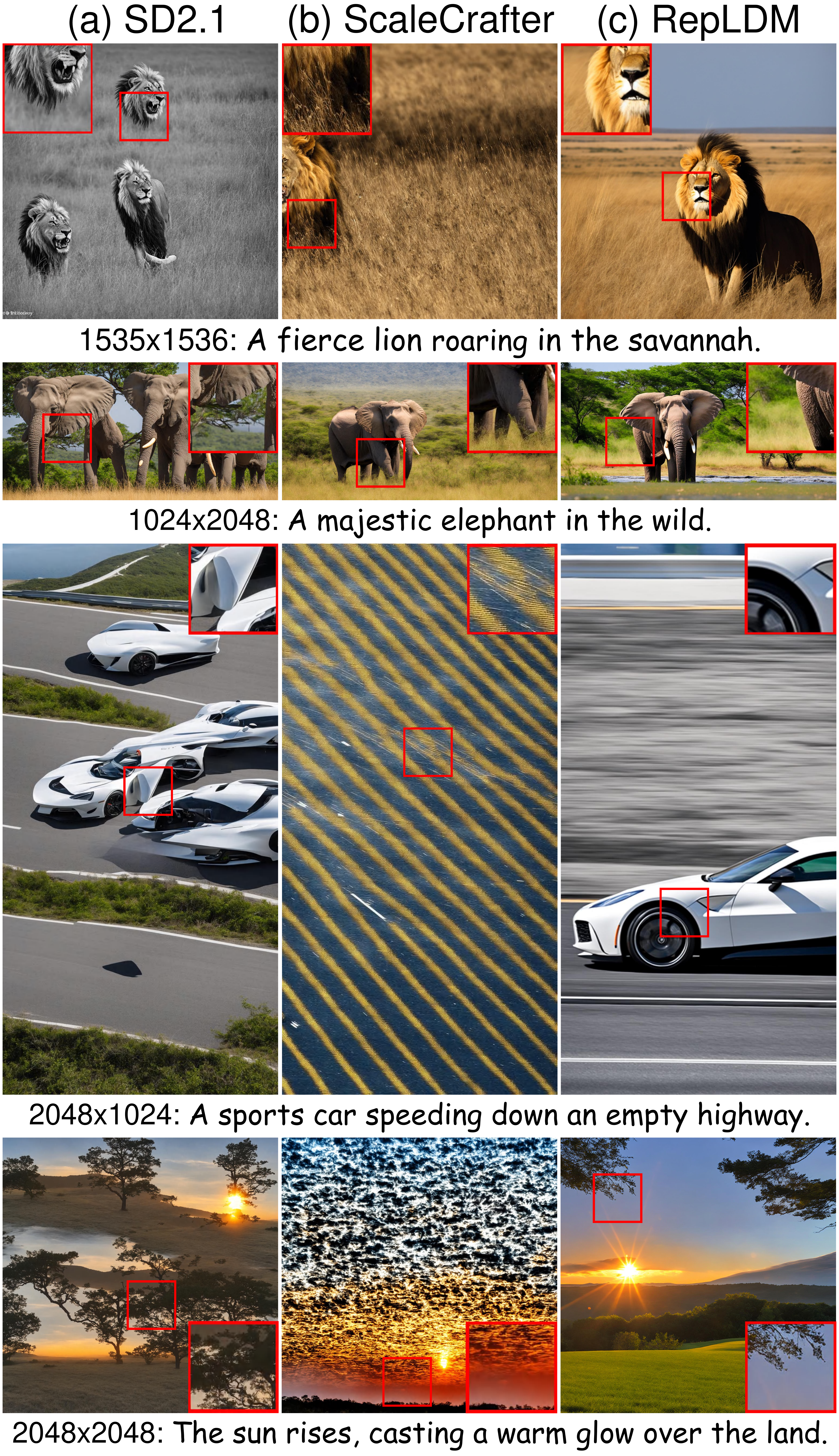}
\vspace{-1.em}
\caption{\textbf{Qualitative comparison using SD2.1 as the pretrained model}.}
\label{fig:comparison_SD2.1}
\vspace{-4em}
\end{wrapfigure}

\paragraph{Qualitative comparison.}
Fig.~\ref{fig:comparison_SD2.1} presents the results of the qualitative comparison. It can be observed that, when generating high-resolution images, SD2.1 also faces issues with repetitive object structures. ScaleCrafter often exhibits structural collapse during denoising with SD2.1, resulting in suboptimal performance. In contrast, \framework consistently produces high-quality results across all resolutions, highlighting the generalizability of the \framework generation framework.

\paragraph{Quantitative comparison.}
Since the code for using SD2.1 as the pretrained model in AccDiffusion and DemoFusion is not publicly available, we compare \framework with ScaleCrafter.
We compared the model performance at four resolutions: $1536\times1536$, $1024\times 2048$, $2048\times 1024$, and $2048\times 2048$.
Considering that SD2.1's generation capabilities are weaker than SDXL, we set \(\eta_2 = [0.2, 0.2, 0.3]\) for the experiments in this section, while keeping other settings consistent with \S\ref{sec:exp}.

Table~\ref{table:appendix_comparison_SD2.1} presents the results of the quantitative comparison, showing that \framework maintains strong performance when using SD2.1 as the pre-trained model. In contrast, ScaleCrafter performs suboptimally, as it tends to produce structural collapse in the generated images, a phenomenon that is more apparent in the qualitative analysis.

\begin{table*}[!th]
\caption{\textbf{Quantitative comparison results based on SD2.1}. The best results are marked in \textbf{bold}.}
\label{table:appendix_comparison_SD2.1}
\centering
\small
\begin{adjustbox}{width=1\linewidth,center}
\begin{tabular}{rrrrrrrrrrrrrrrrrrrrr}
\toprule[1.5pt]
\multirow{2}{*}{\textbf{Method}} & \multicolumn{5}{c|}{$1536 \times 1536$} & \multicolumn{5}{c|}{$1024 \times 2048$} & \multicolumn{5}{c|}{$2048 \times 1024$} & \multicolumn{5}{c}{$2048 \times 2048$} \\
\cline{2-21} 
& FID & IS & FID$_{c}$ & IS$_{c}$ & CLIP
& FID & IS & FID$_{c}$ & IS$_{c}$ & CLIP
& FID & IS & FID$_{c}$ & IS$_{c}$ & CLIP
& FID & IS & FID$_{c}$ & IS$_{c}$ & CLIP\\
\midrule
SD2.1~\citep{sd}
& 95.4 & 17.8 & 83.4 & 15.8 & 25.0
& 85.8 & 15.9 & 76.1 & 16.3 & \textbf{25.2}
& 101.8 & 15.8 & 79.8 & 16.8 & 24.6
& 121.7 & 14.4 & 92.7 & 14.4 & 24.5\\
ScaleCrafter~\cite{he2023scalecrafter}
& 140.4 & 10.6 & 136.4 & 9.7 & 21.9
& 150.0 & 10.1 & 139.3 & 10.1 & 21.7
& 149.8 & 10.4 & 135.6 & 11.5 & 21.8
& 144.2 & 10.4 & 135.2 & 10.3 & 23.4\\
\framework
& \textbf{60.3} & \textbf{21.0} & \textbf{50.6} & \textbf{18.3} & \textbf{25.4}
& \textbf{61.1} & \textbf{19.9} & \textbf{54.1} & \textbf{18.4} & 25.0
& \textbf{63.7} & \textbf{19.2} & \textbf{50.4} & \textbf{18.2} & \textbf{24.7}
& \textbf{60.5} & \textbf{21.5} & \textbf{48.8} & \textbf{17.2} & \textbf{25.3}\\
\bottomrule[1.5pt]
\end{tabular}
\end{adjustbox}
\vspace{-.5em}
\end{table*}

\subsection{Ablation Study on Attention Guidance}

\paragraph{Quantitative ablation.}
Table~\ref{table:ablation_ag_SD2.1} shows the results of the quantitative ablation on \moduleone using SD2.1 as the pretrained model.
It can be observed that \moduleone leads to improvements in metrics. These improvements are more evident in the qualitative ablation analysis.

\begin{table*}[ht!]
\vspace{-.5em}
\caption{\textbf{Quantitative ablation results based on SD2.1}. The best results are marked in \textbf{bold}.}
\label{table:ablation_ag_SD2.1}
\centering
\small
\begin{adjustbox}{width=1\linewidth,center}
\begin{tabular}{rrrrrrrrrrrrrrrrrrrrr}
\toprule[1.5pt]
\multirow{2}{*}{\textbf{Method}} & \multicolumn{5}{c|}{$1536 \times 1536$} & \multicolumn{5}{c|}{$1024 \times 2048$} & \multicolumn{5}{c|}{$2048 \times 1024$} & \multicolumn{5}{c}{$2048 \times 2048$} \\
\cline{2-21} 
& FID & IS & FID$_{c}$ & IS$_{c}$ & CLIP
& FID & IS & FID$_{c}$ & IS$_{c}$ & CLIP
& FID & IS & FID$_{c}$ & IS$_{c}$ & CLIP
& FID & IS & FID$_{c}$ & IS$_{c}$ & CLIP\\
\midrule
w/o \moduleoneshort
& 61.2 & 20.9 & \textbf{50.2} & \textbf{18.9} & 25.2
& 61.5 & 19.6 & \textbf{54.0} & \textbf{19.5} & 24.9
& 64.6 & \textbf{19.6} & \textbf{49.2} & 17.0 & 24.6
& 61.1 & 21.2 & \textbf{46.5} & \textbf{18.2} & 25.2\\
w/ \moduleoneshort
& \textbf{60.3} & \textbf{21.0} & 50.6 & 18.3 & \textbf{25.4}
& \textbf{61.1} & \textbf{19.9} & 54.1 & 18.4 & \textbf{25.0}
& \textbf{63.7} & 19.2 & 50.4 & \textbf{18.2} & \textbf{24.7}
& \textbf{60.5} & \textbf{21.5} & 48.8 & 17.2 & \textbf{25.3}\\
\bottomrule[1.5pt]
\end{tabular}
\end{adjustbox}
\vspace{-.5em}
\end{table*}

\begin{figure}[!th]
\centering
\includegraphics[width=1\textwidth]{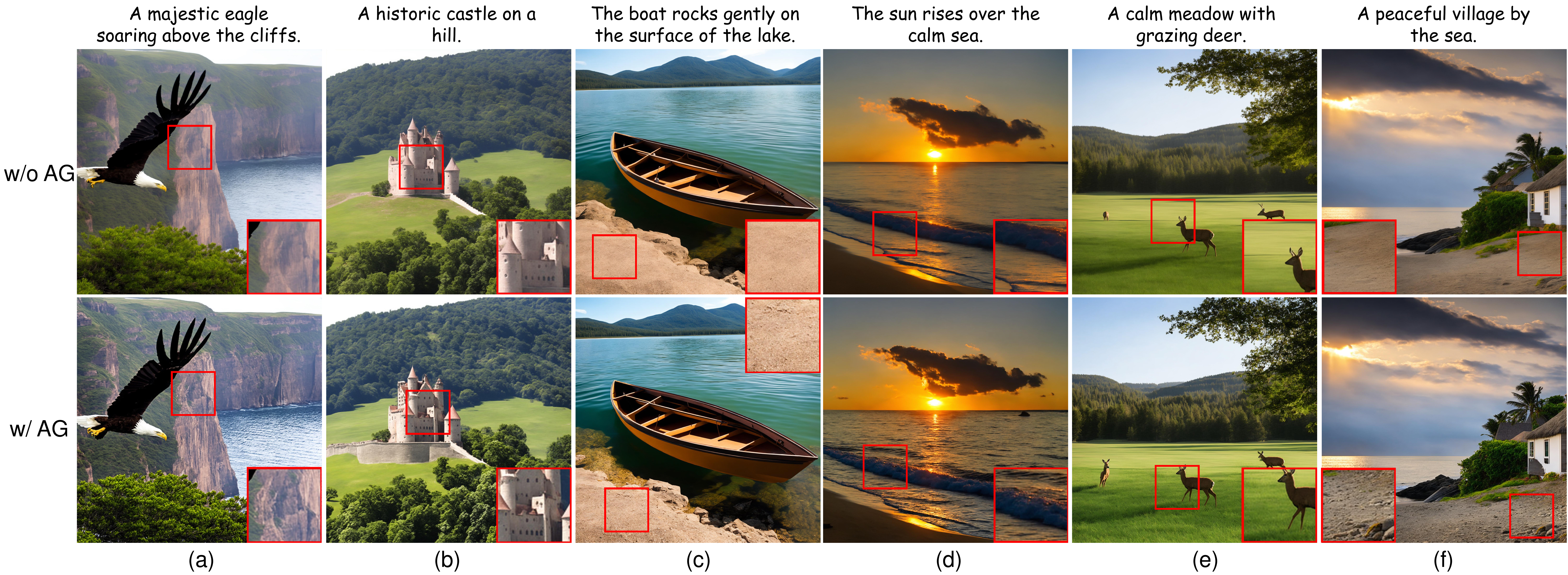}
\vspace{-2em}
\caption{\textbf{Ablation study of \moduleone using SD2.1 as the pre-trained model}. Resolution: $2048\times 2048$.}
\vspace{-1.em}
\label{fig:ag_ablation_SD2.1}
\end{figure}

\paragraph{Qualitative ablation.}
Fig.~\ref{fig:ag_ablation_SD2.1} presents the ablation analysis of \moduleone based on SD2.1. From the figure, it can be observed that \moduleone also enhances detail richness and color vibrancy when using SD2.1, further demonstrating its generalization capability.

\section{Attention Guidance Also Works in Other Generation Frameworks}
\label{apx:ablation_with_other_framework}
In this section, we apply \moduleone to other generative frameworks to demonstrate its generalization capability. Specifically, we apply \moduleone to the generative frameworks of HiDiffusion and DemoFusion, and perform both quantitative and qualitative ablation studies.

\subsection{Quantitative Ablation}
In this section, considering the long inference time of DemoFusion, we perform quantitative ablation studies on \moduleone using the HiDiffusion generation frameworks at a resolution of $2048\times 2048$. All experimental settings are consistent with those in \S\ref{sec:exp}.

\begin{wraptable}{r}{6cm}
\vspace{-2em}
\scriptsize
\caption{\textbf{Quantitative ablation of \moduleone using HiDiffusion frameworks}. The best results are marked in bold.}
\label{table:ag_ablation_additional}
\centering
\begin{adjustbox}{width=0.95\linewidth,center}
\begin{tabular}{rrrrrr}
\toprule[1.5pt]
\textbf{Method} & FID & IS  & FID$_c$ & IS$_c$ & CLIP\\
\midrule
HiDiffusion~\citep{zhang2025hidiffusion}
& 81.0 & 16.8 & 64.1 & 14.2 & \textbf{24.9}\\
HiDiffusion+\moduleoneshort
& \textbf{79.4} & \textbf{17.0} & \textbf{62.4} & \textbf{14.6} & \textbf{24.9}\\
\bottomrule[1.5pt]
\end{tabular}
\end{adjustbox}
\vspace{-.8em}
\end{wraptable}

Table~\ref{table:ag_ablation_additional} presents the quantitative ablation results using the HiDiffusion framework. It is evident that incorporating \moduleone improves HiDiffusion across all metrics. This is further corroborated by the qualitative analysis in Fig.~\ref{fig:ag_for_hidiff}, which demonstrates that \moduleone alleviates some of the structural collapses observed in HiDiffusion.

\subsection{Qualitative Ablation}
\begin{figure*}[t!]
\centering
\includegraphics[width=1\textwidth]{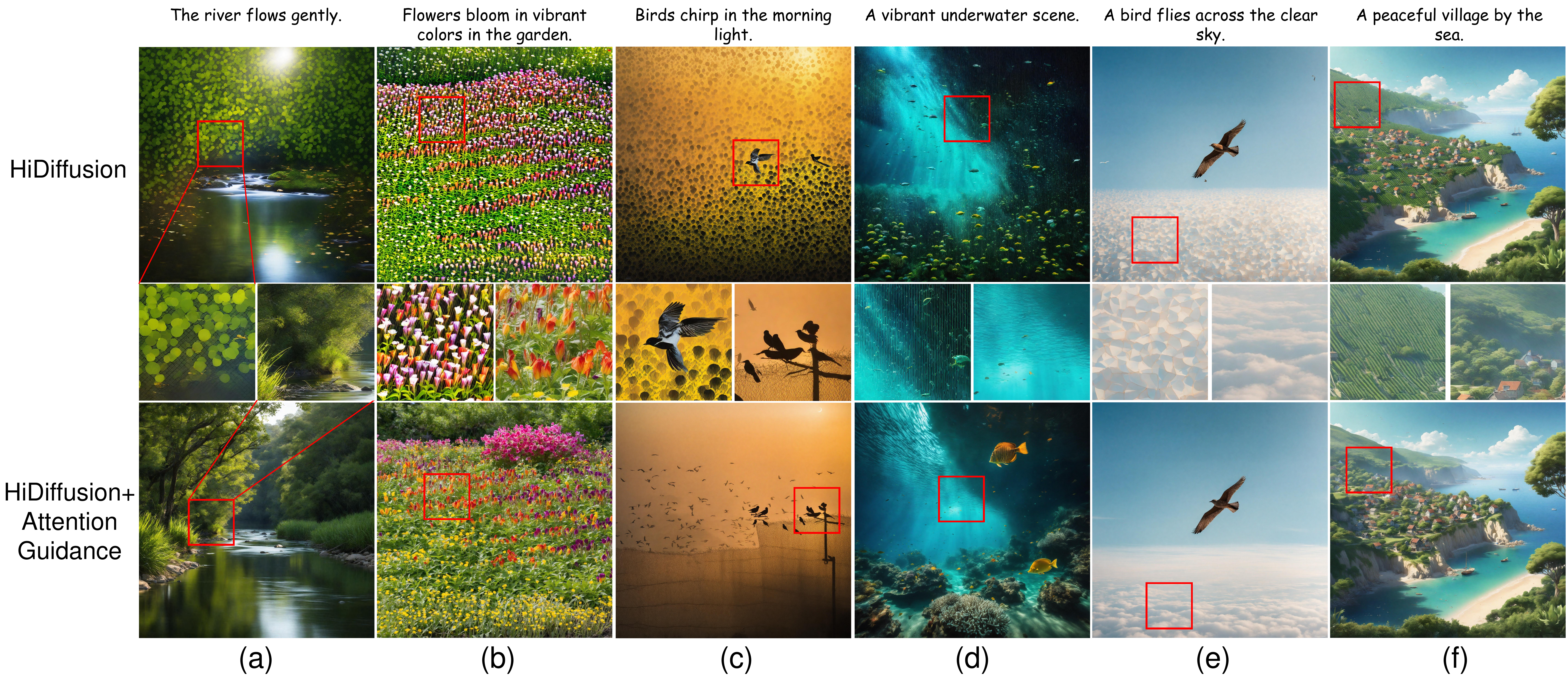}
\vspace{-2em}
\caption{
\textbf{Qualitative ablation of \moduleone in the HiDiffusion Framework}. All images have a resolution of \(2048 \times 2048\).
Figures (a)-(c) demonstrate that \moduleone can mitigate the issue of structural collapse in generated images, while Figures (d)-(f) show that \moduleone resolves structural deformation issues and enhances image details.
}
\label{fig:ag_for_hidiff}
\vspace{-1em}
\end{figure*}

\paragraph{HiDiffusion+\moduleone.}
We incorporate \moduleone into the generative framework of HiDiffusion. Fig.~\ref{fig:ag_for_hidiff}~(a)-(c) demonstrate that using \moduleone effectively mitigates the issue of structural collapse in synthesized images. Fig.~\ref{fig:ag_for_hidiff}~(d)-(f) further show that \moduleone can also address the structural deformation inherent to HiDiffusion, enhance image details, and improve overall image quality.

\begin{figure*}[ht!]
\centering
\includegraphics[width=1\textwidth]{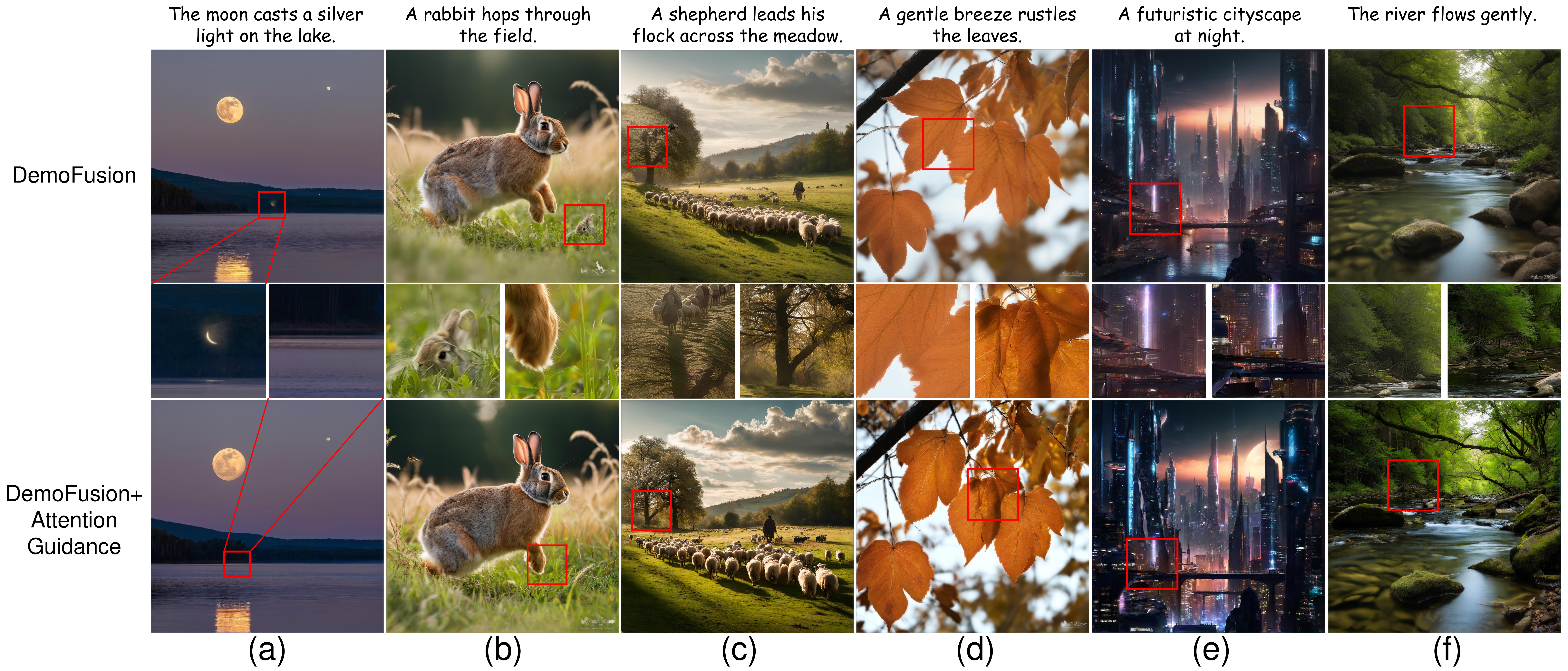}
\vspace{-2em}
\caption{\textbf{Qualitative ablation of \moduleone in the DemoFusion Framework}. All images have a resolution of \(2048 \times 2048\). Figures (a)-(c) demonstrate that \moduleone effectively mitigates the issue of repetitive structures in images, while Figures (d)-(f) showcase \moduleone's ability to enrich image details.}
\label{fig:ag_for_demofusion}
\vspace{-.5em}
\end{figure*}

\paragraph{DemoFusion+\moduleone.}
We incorporate \moduleone into the generative framework of DemoFusion. As shown in Fig.~\ref{fig:ag_for_demofusion}~(a)-(c), \moduleone effectively mitigates the issue of repetitive structures in DemoFusion. Fig.~\ref{fig:ag_for_demofusion}~(d)-(f) further illustrate role of \moduleone in enriching image details and enhancing overall image quality.

\section{Super-Resolved Images Tend to Lack High-Resolution Details}
\label{apx:explain_SR}
To explain why using super-resolution models to obtain HR images is sub-optimal, in this section, we conduct both qualitative and quantitative comparisons between \framework and the super-resolution results.
Specifically, we use BSRGAN~\citep{zhang2021designing} to upsample the generated results of SDXL~\citep{sdxl} at its training resolution.

\paragraph{Quantitative results.} As shown in table.~\ref{table:compared_to_SR}, the super-resolution model (SDXL + BSRGAN) demonstrate comparable performance in quantitative experiments, a phenomenon also observed in the DemoFusion's experiments.
This is because super-resolution models can at least preserve the low-frequency structures of images without significant errors.
However, quantitative metrics such as FID and IS, are widely recognized as insufficient for comprehensively evaluating the performance of model’s generation. As a result, user studies are commonly employed to provide human-level evaluation with more intuition~\citep{lucic2018gans, sd, sdxl, controlnet, guo2023animatediff, he2023scalecrafter, jin2024training}.
For example, in ScaleCrafter~\citep{he2023scalecrafter}, they conducted both quantitative and user study analyses in comparison with the SD+SR approach. Their results show that, although ScaleCrafter performs worse than SD+SR on quantitative metrics, users significantly prefer the textures and details generated by ScaleCrafter. One important reason is that the goal of the SR model is to produce images consistent with the input, which limits its performance in high-resolution generation – needing more detail for true high-resolution visuals beyond simple smoothing~\citep{he2023scalecrafter, du2024demofusion, lin2024accdiffusion, lin2024accdiffusionV2, jin2024training, guo2024make, zheng2024any}.

\begin{table*}[ht!]
\vspace{-.5em}
\caption{\textbf{Quantitative comparison results between \framework and SDXL+BSRGAN}. The best results are marked in \textbf{bold}.}
\label{table:compared_to_SR}
\centering
\small
\begin{adjustbox}{width=1\linewidth,center}
\begin{tabular}{rrrrrrrrrrrrrrrrrrrrr}
\toprule[1.5pt]
\multirow{2}{*}{\textbf{Method}} & \multicolumn{5}{c|}{$2048 \times 2048$} & \multicolumn{5}{c|}{$2048 \times 4096$} & \multicolumn{5}{c|}{$4096 \times 2048$} & \multicolumn{5}{c}{$4096 \times 4096$} \\
\cline{2-21} 
& FID & IS & FID$_{c}$ & IS$_{c}$ & CLIP
& FID & IS & FID$_{c}$ & IS$_{c}$ & CLIP
& FID & IS & FID$_{c}$ & IS$_{c}$ & CLIP
& FID & IS & FID$_{c}$ & IS$_{c}$ & CLIP\\
\midrule
SDXL+BSRGAN
& 66.2 & 21.1 & 47.5 & 16.6 & \textbf{25.7}
& \textbf{80.7} & 19.8 & \textbf{50.2} & 12.3 & \textbf{25.1}
& \textbf{92.7} & 17.6 & 57.9 & 12.1 & \textbf{24.9}
& \textbf{90.0} & 20.9 & \textbf{56.0} & 13.8 & \textbf{25.2}\\
\framework
& \textbf{66.0} & 21.0 & \textbf{47.4} & 17.5 & 25.1
& 89.0 & \textbf{20.3} & 56.0 & \textbf{19.0} & 25.0
& 93.2 & \textbf{19.5} & \textbf{56.9} & \textbf{16.5} & \textbf{24.9}
& 90.6 & \textbf{21.1} & 59.0 & \textbf{14.8} & 24.6\\
\bottomrule[1.5pt]
\end{tabular}
\end{adjustbox}
\vspace{-.5em}
\end{table*}

\begin{figure*}[t!]
\centering
\includegraphics[width=1\textwidth]{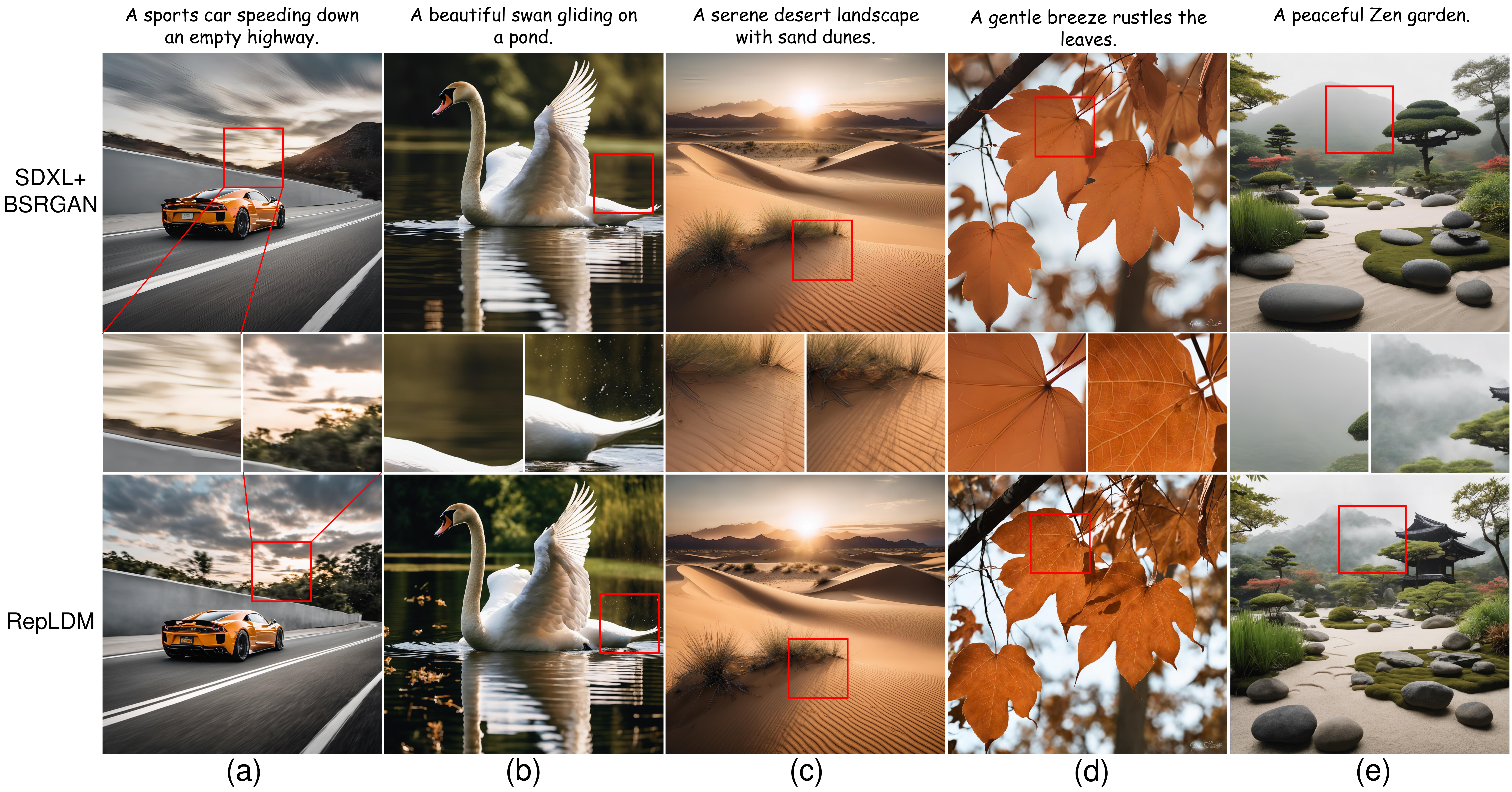}
\vspace{-2em}
\caption{\textbf{Qualitative comparison with SDXL+BSRGAN}. The prompts for the generated images are provided above the figures. The resolution of (a) to (c) are $2048\times 2048$, and the resolution of (d) and (e) are $4096\times 4096$.}
\label{fig:comparison_with_sr}
\vspace{-1.5em}
\end{figure*}

\paragraph{Qualitative results.} As shown in Fig.~\ref{fig:comparison_with_sr}, compared to \framework, SDXL+BSRGAN, while maintaining decent image structure, fails to generate the level of detail expected from HR images. The absence of these details sometimes leads to the model's inability to simulate realistic scenes.
For example, in Fig.~\ref{fig:comparison_with_sr}~(c), SDXL+BSRGAN fails to generate realistic shadows.

\section{Memory Usage Analysis}
\label{apx:memory_analyze}
We compare the GPU memory usage required by the models.
Specifically, we test the minimum GPU memory requirements during model inference based on the model's open-source code.
Table~\ref{table:appendix_model_memory_usage} shows the resource consumption of different models when generating images at various resolutions.

\begin{table}[ht!]
\vspace{-0.3em}
\caption{\textbf{Model Memory Usage (GB).} The best results are marked in \textbf{bold}, and the second best results are marked by \underline{underline}.}
\vspace{-.5em}
\centering
\scriptsize
\begin{tabular}{rrrr}
\toprule[1.5pt]
\textbf{Resolutions} & $2048\times 2048$ & $2048\times 4096$ & $4096\times 4096$\\
\midrule
SDXL~\citep{sdxl}
& \underline{15.9} & \textbf{16.1} & \underline{16.6}\\
MultiDiff.~\citep{bar2023multidiffusion}
& 22.0 & 16.8 & 16.8\\
ScaleCrafter~\citep{he2023scalecrafter}
& 17.4 & 17.6 & 19.1\\
UG~\citep{hwang2024upsample}
& 23.9 & 16.5 & 18.0\\
DemoFusion~\citep{du2024demofusion}
& \textbf{15.2} & 18.4 & 16.8\\
AccDiff.~\citep{lin2024accdiffusion}
& 22.1 & 23.0 & 22.1\\
HiDiff.~\citep{zhang2025hidiffusion}
& 23.9 & \underline{16.2} & \textbf{16.2}\\
\framework
& 16.0 & 21.1 & 23.8\\
\bottomrule[1.5pt]
\end{tabular}
\label{table:appendix_model_memory_usage}
\vspace{-2.em}
\end{table}

It is worth noting that for HR image generation tasks, the memory bottleneck lies in the encoding and decoding of the VAE rather than interpolating the image in pixel space.
To address the challenges of encoding and decoding HR images, researchers typically employ tiled encoders and tiled decoders.
In this work, we also utilize a tiled-encoder and decoder when generating ultra-high-resolution images, allowing us to generate images with resolutions up to \(4096 \times 7280\) or higher on a 24GB VRAM NVIDIA 3090 GPU (as shown in Fig.~\ref{fig:teaser}).

It is important to note that different models have undergone varying degrees of additional optimization in their official open-source implementations.
Specifically, some open-source codes utilize existing optimization tools, such as accelerate~\citep{accelerate} or Flash Attention~\citep{dao2022flashattention}, which provide additional advantages in terms of inference speed and memory usage performance.
To ensure a fair comparison, in Table~\ref{table:appendix_model_memory_usage}, we did not use such additional optimizations in the implementation of \framework.


\clearpage

\newpage
\section*{NeurIPS Paper Checklist}

\begin{enumerate}

\item {\bf Claims}
    \item[] Question: Do the main claims made in the abstract and introduction accurately reflect the paper's contributions and scope?
    \item[] Answer: \answerYes{} 
    \item[] Justification: The main claims reflect the paper’s contributions and scope.

\item {\bf Limitations}
    \item[] Question: Does the paper discuss the limitations of the work performed by the authors?
    \item[] Answer: \answerYes{} 
    \item[] Justification: We discuss the limitations in \S\ref{sec:limitations}, analyze their underlying causes, and discuss potential directions for future work.

\item {\bf Theory assumptions and proofs}
    \item[] Question: For each theoretical result, does the paper provide the full set of assumptions and a complete (and correct) proof?
    \item[] Answer: \answerNA{} 
    \item[] Justification: The paper does not include theoretical results.

    \item {\bf Experimental result reproducibility}
    \item[] Question: Does the paper fully disclose all the information needed to reproduce the main experimental results of the paper to the extent that it affects the main claims and/or conclusions of the paper (regardless of whether the code and data are provided or not)?
    \item[] Answer: \answerYes{} 
    \item[] Justification: We provide a detailed description of the methodology and its underlying ideas in \S\ref{sec:method}. In addition, we present the full algorithmic pipeline using pseudocode in Appendix~\ref{apx:framework_algorithm}. We also commit to open-sourcing the code of our method.

\item {\bf Open access to data and code}
    \item[] Question: Does the paper provide open access to the data and code, with sufficient instructions to faithfully reproduce the main experimental results, as described in supplemental material?
    \item[] Answer: \answerYes{} 
    \item[] Justification: In Appendix~\ref{apx:framework_algorithm}, we detail the method using pseudocode. Additionally, upon acceptance, we will clean and release our code base and share it on GitHub. All data used in this paper belongs to existing open source datasets and have been correctly cited to ensure reproduction.

\item {\bf Experimental setting/details}
    \item[] Question: Does the paper specify all the training and test details (e.g., data splits, hyperparameters, how they were chosen, type of optimizer, etc.) necessary to understand the results?
    \item[] Answer: \answerYes{} 
    \item[] Justification: The proposed method requires no training. We provide a detailed explanation of the inference and evaluation settings in \S\ref{sec:exp_implementation}. We determine the hyperparameters through ablation studies, with details provided in Appendix~\ref{apx:supplementary_ablation}.

\item {\bf Experiment statistical significance}
    \item[] Question: Does the paper report error bars suitably and correctly defined or other appropriate information about the statistical significance of the experiments?
    \item[] Answer: \answerYes{} 
    \item[] Justification: We repeated the experiments in Table~\ref{table:comparison} in Appendix~\ref{apx:robustness_analysis}, computing the mean and standard deviation, and also conducted additional replication experiments on the LAION dataset~\cite{schuhmann2022laion}.

\item {\bf Experiments compute resources}
    \item[] Question: For each experiment, does the paper provide sufficient information on the computer resources (type of compute workers, memory, time of execution) needed to reproduce the experiments?
    \item[] Answer: \answerYes{} 
    \item[] Justification: We report the resources needed to reproduce the experiments in \S\ref{sec:exp_implementation}.
    
\item {\bf Code of ethics}
    \item[] Question: Does the research conducted in the paper conform, in every respect, with the NeurIPS Code of Ethics \url{https://neurips.cc/public/EthicsGuidelines}?
    \item[] Answer: \answerYes{} 
    \item[] Justification: We conform to the NeurIPS code of ethics.

\item {\bf Broader impacts}
    \item[] Question: Does the paper discuss both potential positive societal impacts and negative societal impacts of the work performed?
    \item[] Answer: \answerNA{} 
    \item[] Justification: the proposed method builds upon a pretrained generative model to produce higher-resolution images in a training-free manner, and thus does not introduce any additional or specific societal impacts.
    
\item {\bf Safeguards}
    \item[] Question: Does the paper describe safeguards that have been put in place for responsible release of data or models that have a high risk for misuse (e.g., pretrained language models, image generators, or scraped datasets)?
    \item[] Answer: \answerNA{} 
    \item[] Justification: Our method does not rely on any specific publicly released model and requires no specialized fine-tuning, and therefore does not necessitate additional safeguards.

\item {\bf Licenses for existing assets}
    \item[] Question: Are the creators or original owners of assets (e.g., code, data, models), used in the paper, properly credited and are the license and terms of use explicitly mentioned and properly respected?
    \item[] Answer: \answerYes{}{} 
    \item[] Justification: We cite all used resources such as implementations of baselines and data. We release our work with CC-By 4.0 license.

\item {\bf New assets}
    \item[] Question: Are new assets introduced in the paper well documented and is the documentation provided alongside the assets?
    \item[] Answer: \answerNA{} 
    \item[] Justification: This study does not involve the release of any new assets.

\item {\bf Crowdsourcing and research with human subjects}
    \item[] Question: For crowdsourcing experiments and research with human subjects, does the paper include the full text of instructions given to participants and screenshots, if applicable, as well as details about compensation (if any)? 
    \item[] Answer: \answerYes{} 
    \item[] Justification: In the user study conducted in this work, we recruited volunteers to evaluate the quality of the generated images. Detailed instructions were provided to the participants. As the participants were volunteers, no compensation was involved.

\item {\bf Institutional review board (IRB) approvals or equivalent for research with human subjects}
    \item[] Question: Does the paper describe potential risks incurred by study participants, whether such risks were disclosed to the subjects, and whether Institutional Review Board (IRB) approvals (or an equivalent approval/review based on the requirements of your country or institution) were obtained?
    \item[] Answer: \answerYes{} 
    \item[] Justification: This study only required volunteers to evaluate the quality of generated images, and therefore poses no particular potential risks. Furthermore, to protect the privacy of participants' preferences, all responses were anonymized and randomized, and no personal information was collected.

\item {\bf Declaration of LLM usage}
    \item[] Question: Does the paper describe the usage of LLMs if it is an important, original, or non-standard component of the core methods in this research? Note that if the LLM is used only for writing, editing, or formatting purposes and does not impact the core methodology, scientific rigorousness, or originality of the research, declaration is not required.
    \item[] Answer: \answerNA{} 
    \item[] Justification: The core method development in this research does not involve LLMs as any important, original, or non-standard components.

\end{enumerate}

\end{document}